\pdfoutput=1
\documentclass[11pt]{article}

\usepackage[]{ACL2023}

\usepackage{times}
\usepackage{latexsym}

\usepackage[T1]{fontenc}
\usepackage[utf8]{inputenc}

\usepackage{microtype}

\usepackage{inconsolata}
\usepackage{amssymb}
\usepackage{pdfpages}
\usepackage{hyperref}
\usepackage{booktabs}
\usepackage{caption}
\usepackage{url}
\usepackage{amsmath,bm}

\title{Efficient and Training-Free Control of Language Generation}

\author{Shangda Wu$^{\dagger}$, Maosong Sun$^{\dagger \ddagger}$\\
$^{\dagger}$ Department of Music AI and Information Technology, Central Conservatory of Music\\
$^{\ddagger}$ Department of Computer Science and Technology, Tsinghua University \\
\texttt{shangda@mail.ccom.edu.cn, sms@tsinghua.edu.cn} \\
}

\begin{document}
\maketitle
\begin{abstract}
In recent years, there has been a growing interest in the development of language models capable of generating text with controllable attributes. While several approaches have been proposed, many of these methods require condition-specific data or significant computational resources. In this study, we propose a novel method called \textsc{Gamma Sampling}, which enables controllable language generation without the need for any training data and maintains a fast generation speed. \textsc{Gamma Sampling} incorporates attribute-related information into the sampling process, effectively guiding the language model to produce text with desired attributes. Our experimental results demonstrate that \textsc{Gamma Sampling}, when applied to GPT2, outperforms representative baselines in terms of diversity, attribute relevance, and overall quality of the generated samples.
\end{abstract}

\section{Introduction}

The emergence of large language models (LMs) has been facilitated by the use of large-scale text data obtained through web crawling. However, while these models can produce high-quality texts, there is limited control over the attributes of the generated output. This poses a significant challenge when utilizing unconditional LMs in scenarios that demand precise control over the generated text. Therefore, the field of controllable language generation, which provides greater control over the output of LMs, has become increasingly relevant in practical applications.

Controllable language generation has made significant strides in recent years (\citealt{weng2021conditional}). However, it remains an open question what the ideal method for controlling the attributes of the generated language is (\citealt{DBLP:journals/corr/abs-2203-13299}, \citealt{DBLP:conf/naacl/YangK21}, \citealt{DBLP:journals/corr/abs-2205-14219}). While training a conditional LM from scratch (\citealt{DBLP:journals/corr/abs-1909-05858}) or fine-tuning an existing LM (\citealt{DBLP:journals/corr/abs-1909-08593}, \citealt{DBLP:conf/emnlp/XuLZTCHH21}) have shown promising results, these approaches often require condition-specific data, making them less applicable to unconditional LMs. Alternatively, data-free methods have been proposed (\citealt{DBLP:conf/emnlp/ShinRLWS20}, \citealt{DBLP:conf/kdd/ZouYZYYT21}, \citealt{DBLP:conf/acl/GhazvininejadSP17}, \citealt{DBLP:journals/corr/abs-2012-15416}, \citealt{DBLP:journals/corr/abs-2112-08726}), but they may have limitations in terms of steerability or computational requirements.

This paper introduces a novel approach, called \textsc{Gamma Sampling}, that allows for the efficient and training-free control of LMs. The development of \textsc{Gamma Sampling} is inspired by the concept of gamma correction (\citealt{applebaum1952gamma}), which is a non-linear operation commonly used for encoding and decoding luminance in video or still image systems. \textsc{Gamma Sampling} is founded on the principle that certain attributes of the generated text are closely associated with the presence of specific tokens, and that by adjusting the probability of these attribute-related tokens, the attributes of the generated text can be controlled. Our main contributions are as follows:

\begin{figure*}[t]
	\centering
        \includegraphics[width=0.93\linewidth]{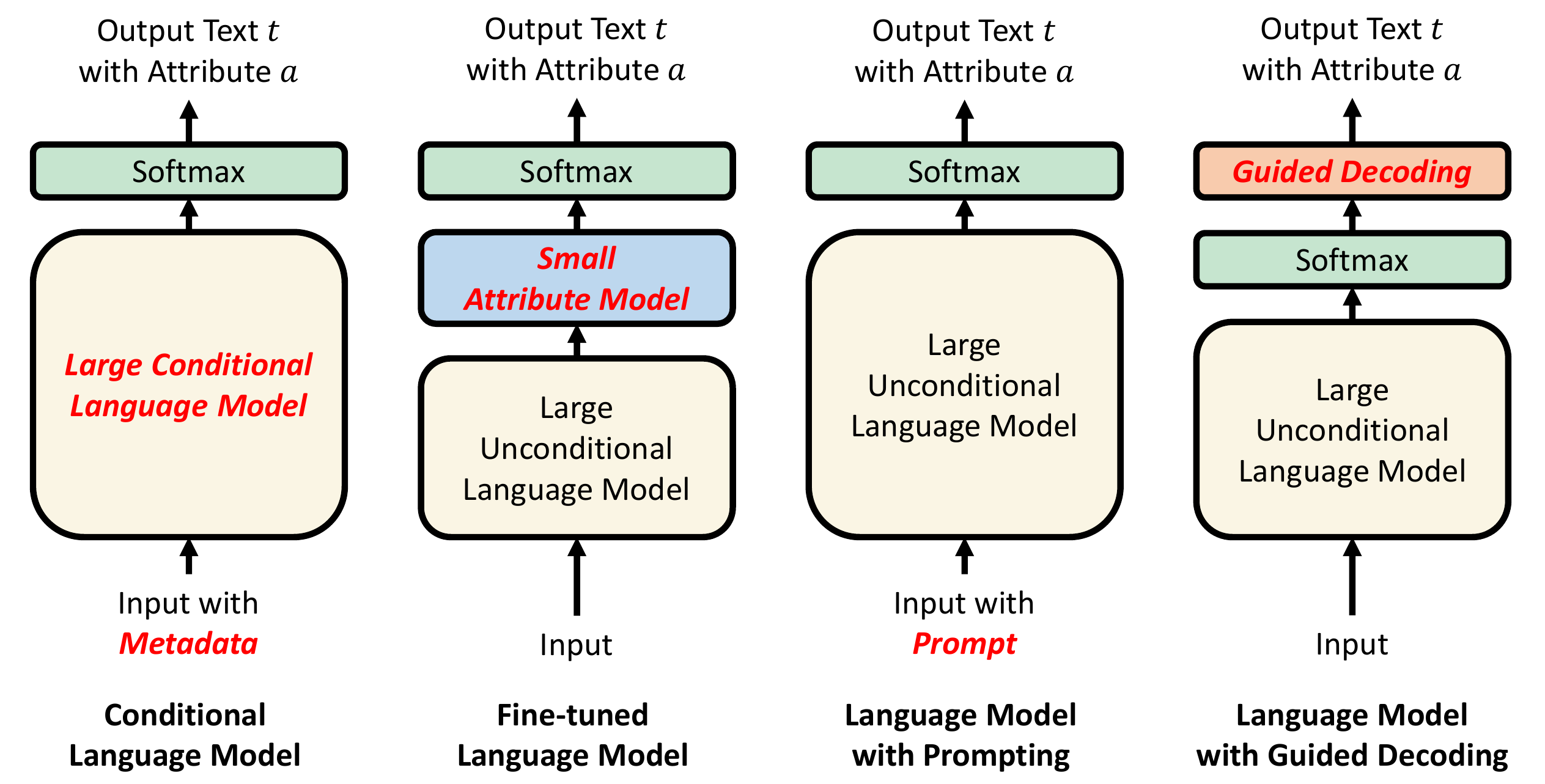}
    \centering
	\caption{Common approaches for controllable language generation. All of them include modules that incorporate attribute-related information, as denoted by the red colouration.} 
    \label{fig1}
\end{figure*}

\begin{itemize}
    \item
    \textsc{Gamma Sampling} is a novel approach for controlling LMs that do not require any training data, and therefore can be readily applied to any LM.
    
    \item
    The method offers the ability to control multiple attributes with fine-grained control strength to users, who can determine the desired level of attribute relevance, and can also turn off the control at any time.
    
    \item
    As \textsc{Gamma Sampling} requires minimal computation at the decoding time, its time cost remains stable, regardless of the scale of the input. Empirical analysis has revealed that \textsc{Gamma Sampling} is significantly faster than other baselines, such as PPLM \citep{DBLP:conf/iclr/DathathriMLHFMY20}, with a generation time that is at least 100 $\times$ faster.

    \item 
     The performance of \textsc{Gamma Sampling}-steered GPT2 was assessed through both automatic and human evaluations, and compared against various other commonly used methods. Results indicate that the proposed method generally outperforms all baselines in terms of diversity, attribute relevance, and overall quality, across all sizes of GPT2 (i.e. Small, Medium, Large, and XL).
\end{itemize}

\section{Background}

The controllable language generation involves the modelling of a probabilistic model $p(t|a)$, which entails generating text $t$, based on a specified attribute $a$. A variety of approaches exist for achieving controllable language generation, each with its advantages and limitations.

\textbf{Conditional Language Model (CLM)} \quad CLMs are a type of LM that is designed to generate text based on certain conditions derived from metadata. However, a large amount of condition-specific data is necessary to train a CLM effectively, and the computational cost of training the entire model from scratch can be significant, making it a resource-intensive task. Additionally, CLMs such as CTRL (\citealt{DBLP:journals/corr/abs-1909-05858}) struggle in controlling the generation of undesirable output, such as detoxification and anti-degeneration (\citealt{DBLP:conf/emnlp/GehmanGSCS20}, \citealt{DBLP:conf/emnlp/MaSRC20}).

\textbf{Fine-tuned Language Model (FTLMs)} \quad FTLMs are an effective solution for balancing training costs and generation quality, as demonstrated by recent research (\citealt{DBLP:journals/corr/abs-1909-08593}, \citealt{DBLP:conf/emnlp/XuLZTCHH21}). These models rely on pre-existing large LMs, with the weights fine-tuned for specific tasks or attributes. Fine-tuning can be limited to the top or additional layers, or introduced through discriminators (\citealt{DBLP:conf/emnlp/KrauseGMKJSR21}, \citealt{DBLP:conf/acl/LiuSLSBSC20}). However, it is important to note that FTLMs still require condition-specific data. Furthermore, while models such as PPLM (\citealt{DBLP:conf/iclr/DathathriMLHFMY20}) that involve combining multiple small attribute models with a large LM have shown promising results, they may lead to decreased computational efficiency due to the need for multiple passes at every decoding step.

\textbf{Prompting} \quad Prompt engineering (\citealt{DBLP:conf/emnlp/ShinRLWS20}, \citealt{DBLP:conf/kdd/ZouYZYYT21}) has become an increasingly prominent training-free approach. By providing carefully crafted prompts, prompt engineering directs large LMs (e.g., ChatGPT\footnote{\url{https://chat.openai.com/}}) towards solving specific tasks and generating more accurate and coherent outputs. However, it is noteworthy that the results of these tasks are significantly influenced by small variations in the prompts (\citealt{DBLP:journals/corr/abs-2205-11916}).

\textbf{Guided Decoding} \quad Guided decoding is a method that can be used to incorporate attribute-related information into the score function for language generation at the decoding time, allowing for the modification of candidate ranking scores. This method has been explored in several studies (\citealt{DBLP:conf/acl/GhazvininejadSP17}, \citealt{DBLP:conf/acl/GhoshCLMS17}, \citealt{DBLP:journals/corr/abs-2012-15416}, \citealt{DBLP:journals/corr/abs-2112-08726}, \citealt{DBLP:conf/acl/LiuSLSBSC20}, \citealt{DBLP:conf/emnlp/PascualEMCW21}). However, a significant challenge associated with guided decoding is the off-distribution issues when control strengths are too strong. Another drawback is the increase in time cost with the number of controlled tokens, which can result in long inference times when there are many controlled tokens. In contrast, Section 3.2 presents the proposed method, \textsc{Gamma Sampling}, which can mitigate or eliminate these disadvantages.

As illustrated in Fig. \ref{fig1}, it is evident that to achieve controllable generation, LMs must be supplied with attribute-related information. The first two approaches rely on extensive training of LMs to acquire attribute-related information from vast amounts of data. Conversely, the latter two believe that large LMs have already acquired sufficient information from data and only require the introduction of attribute-related information into the input or decoding to be implemented for various downstream tasks.

\section{Methodology}
This section provides an overview of gamma correction in Section 3.1, and goes on to present an in-depth analysis of the technical aspects of \textsc{Gamma Sampling} in Section 3.2. Finally, Section 3.3 introduces the practical application of \textsc{Gamma Sampling} to define various controllable attributes.

\subsection{Gamma Correction}

Gamma correction (\citealt{applebaum1952gamma}) is a well-known technique that was initially to compensate for the non-linearity of the input-output characteristic of Cathode Ray Tube (CRT) displays. Since then, this technique has been widely used for a variety of purposes, including adjusting the saturation of images. This approach applies a power function to all pixels of the input image, as the human perception of brightness is known to be non-linear (\citealt{neri2009nonlinear}). In its simplest form, this correction technique is mathematically expressed as a power-law function:

\begin{equation}
\begin{aligned}
     V_{out}=AV_{in}^{\gamma},
     \end{aligned}
\end{equation}

\noindent
where the luminance value $V_{in} \in [0,1]$ is raised to the power $\gamma \in [0, +\infty)$ and multiplied by the constant $A$ (in the common case, $A = 1$) to get the output value $V_{out} \in [0,1]$. 

\subsection{\textsc{Gamma Sampling}}
\subsubsection{Basic Principles}

In linguistic research, it has been observed that the frequency of any word in common corpora is inversely proportional to its ranking in the frequency table (\citealt{DBLP:conf/conll/Powers98}). Given this observation, it is reasonable to employ non-linear processes for tuning the probabilities of attribute-related tokens. Temperature sampling (\citealt{DBLP:conf/mtsummit/DabreF21}), as one example, is also based on a non-linear operation to scale the probability distribution.

\begin{figure}[t]
    \centering
    \begin{minipage}[t]{0.45\textwidth}
        \centering
        \includegraphics[width=0.8\linewidth]{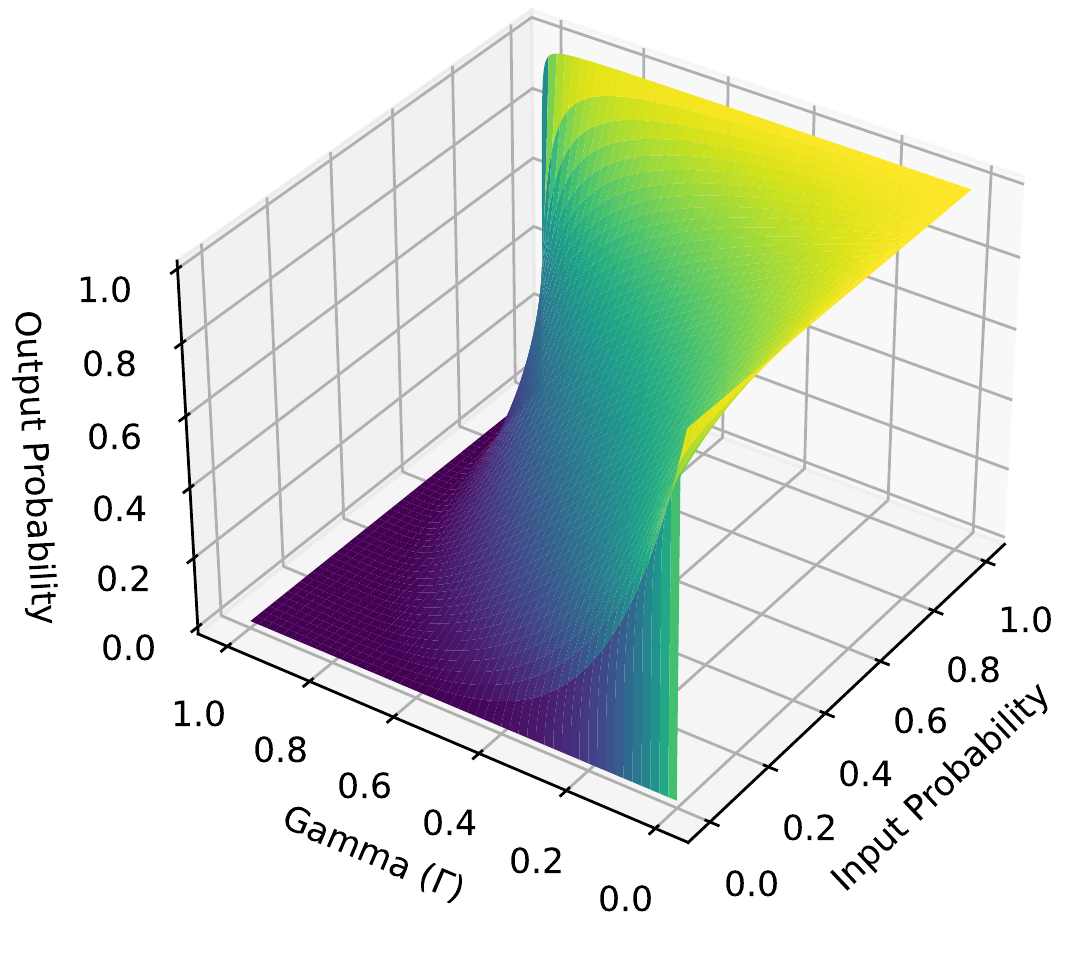}
        \\ (a) 3D surface plot of \textsc{Gamma Sampling}
        \vspace{1em}
    \end{minipage}
    \vspace{1em}
    \begin{minipage}[t]{0.45\textwidth}
        \centering
        \includegraphics[width=0.8\linewidth]{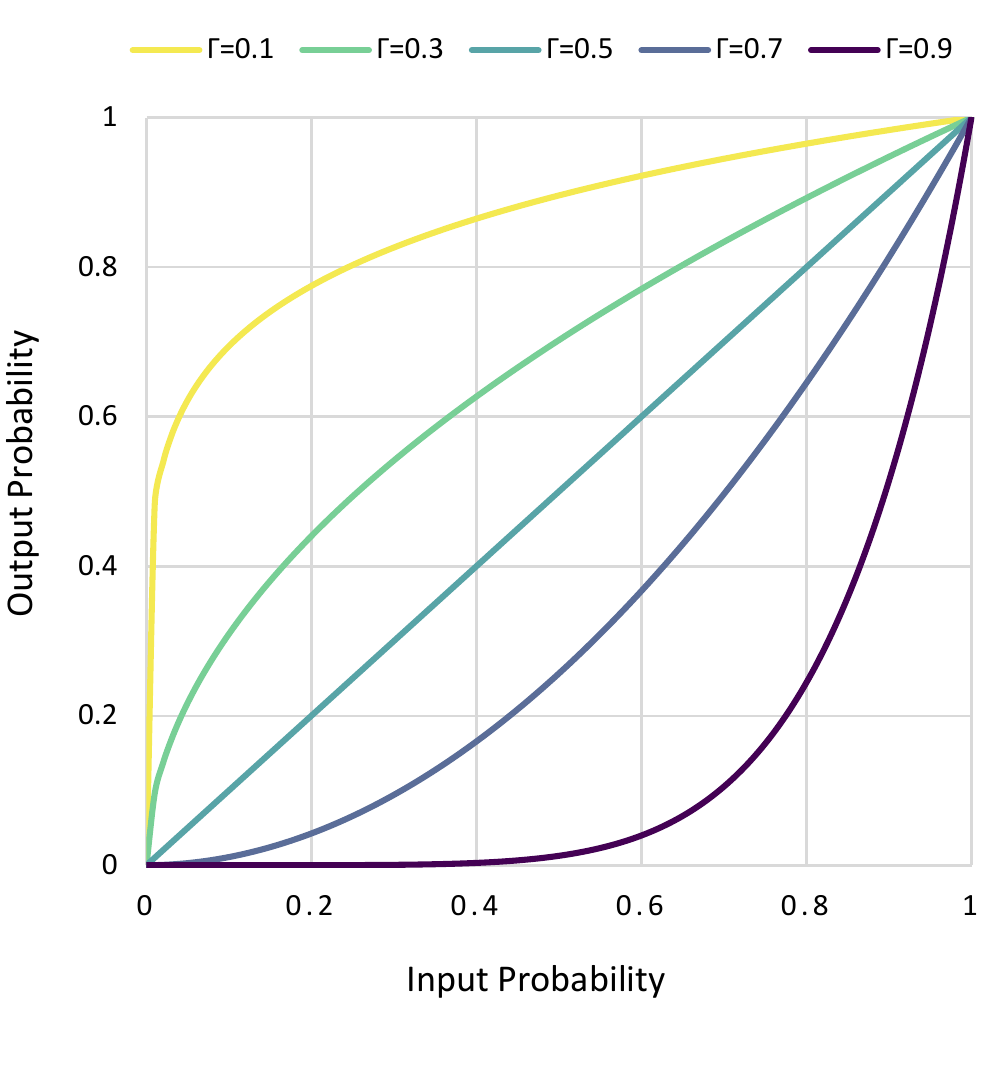}
        \vspace{-1em}
        \\ (b) \textsc{Gamma Sampling} with different $\Gamma$
        \vspace{-1em}
    \end{minipage}
    \caption{The non-linear scaling for the input probabilities by \textsc{Gamma Sampling}. When $\Gamma<0.5$, the probabilities are increased. Conversely, when $\Gamma>0.5$, the probabilities are decreased.}
    \vspace{-1em}
    \label{fig2}
\end{figure}

The non-linear nature of languages indicates that gamma correction may also be used to control text attributes. In this context, the manipulation of text attributes is referred to as \textsc{Gamma Sampling}, with its fundamental operation formalized as follows:

\begin{equation}
\begin{aligned}
    p_{\mathcal{A}_{out}}&=p_{\mathcal{A}_{in}}^{tan(\frac{\pi \Gamma}{2})},
\end{aligned}
\end{equation}

\noindent
where $p_{\mathcal{A}_{in}}$ is the sum of the input probabilities of all tokens in the attribute-related token set $\mathcal{A}$, $p_{\mathcal{A}_{out}}$ is the output one, and $\Gamma \in [0, 1]$ is the control strength. As shown in Fig. \ref{fig2}, the control strength $\Gamma$ determines the degree to which the probabilities of these tokens increase or decrease, and the increase and decrease in probabilities occur over equal length ranges. When $\Gamma=0.5$, there is no change in the probability distribution, while when $\Gamma<0.5$, the probabilities of the attribute-related tokens increase and vice versa.

It is important to consider the potential risks of using \textsc{Gamma Sampling} on its own. When setting $\Gamma$ to a very low level, even tokens with low probability may experience a significant increase in their probabilities, which may result in off-distribution. Therefore, it is advisable to use either top-$k$ sampling \cite{DBLP:conf/acl/LewisDF18} or nucleus sampling \cite{DBLP:conf/iclr/HoltzmanBDFC20} before \textsc{Gamma Sampling} to eliminate these unreliable attribute-related tokens.

\textsc{Gamma Sampling} is an innovative approach that sets it apart from other methods. Notably, it operates on probability distributions rather than the logits (\citealt{DBLP:conf/acl/LiuSLSBSC20}), allowing for top-$k$ or nucleus sampling to be applied before it, thereby avoiding over-control strengths that may result in off-distribution. Additionally, \textsc{Gamma Sampling} eliminates the need for customization of the score function (\citealt{DBLP:conf/acl/GhazvininejadSP17}), additional discriminators (\citealt{DBLP:conf/emnlp/KrauseGMKJSR21}), or the use of specific external models (\citealt{DBLP:conf/emnlp/PascualEMCW21}). This method requires only minimal computation at the decoding time, and therefore is computationally efficient and independent of the number of controlled tokens, resulting in a nearly constant time cost.

\begin{figure}[t]
	\centering
        \includegraphics[width=0.8\linewidth]{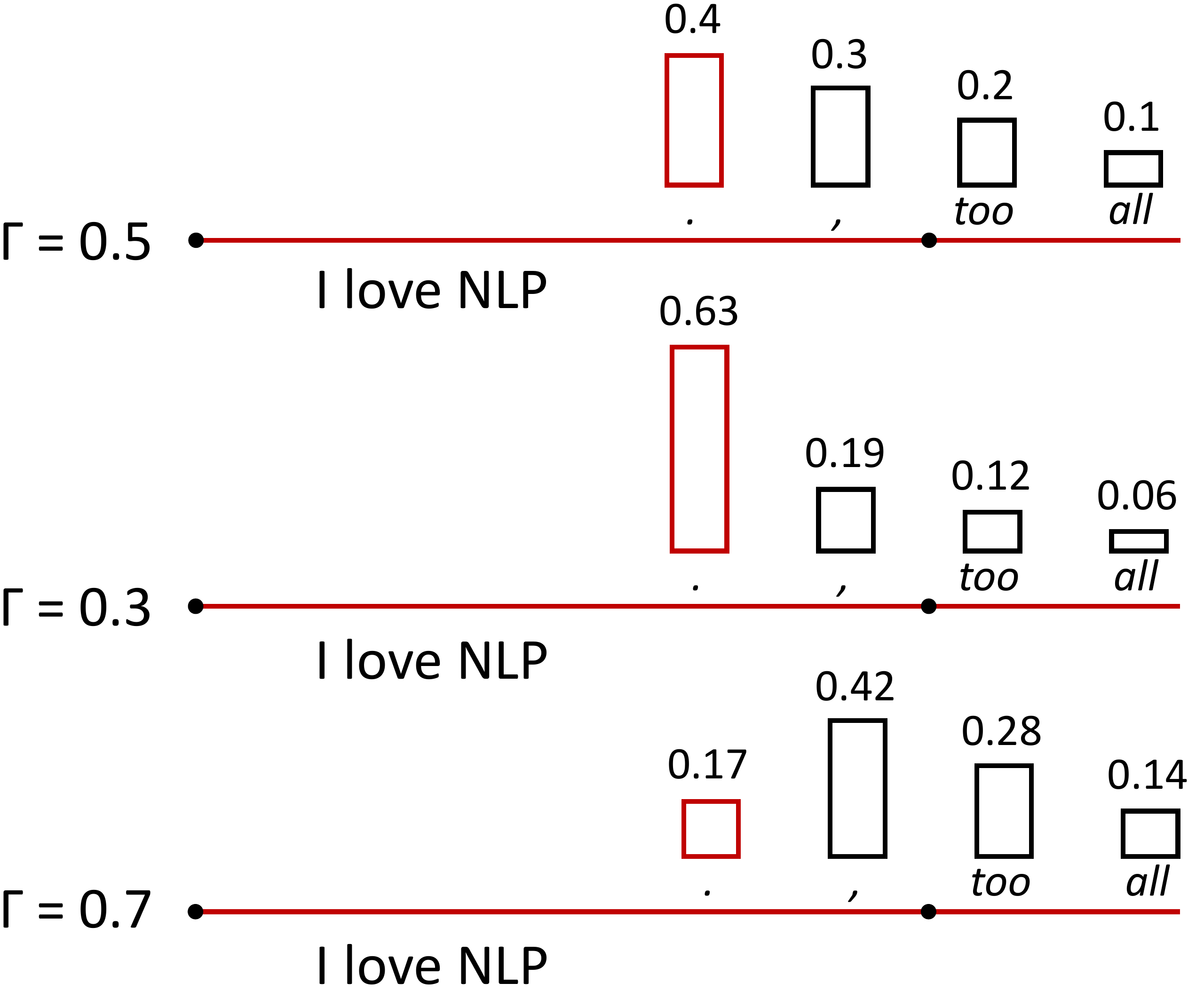}
    \centering
	\caption{An example of controllable sentence length with the input \texttt{I love NLP}, where the full stop is the attribute-related token. See Section 4.2 for more details.}
	\label{fig3}
    \vspace{-1em}
\end{figure}

\subsubsection{Controlling Single Attribute}

The use of \textsc{Gamma Sampling} in fine-grained control involves the scaling of probabilities associated with attribute-related tokens. To achieve optimal steerability, certain post-processing steps must be taken, which include the ability to scale the sum of probabilities of attribute-related tokens to 0 or 1, maintaining the ratio of probabilities between tokens of the same type (i.e., attribute-related or non-attribute-related), and ensuring the validity of the output probability distribution. The process of controlling a single attribute using \textsc{Gamma Sampling} can be formalized as follows:

\begin{equation}
\resizebox{.85\hsize}{!}{$
\begin{aligned}
    p_{\mathcal{A}_{out}}&=p_{\mathcal{A}_{in}}^{tan(\frac{\pi \Gamma}{2})}, \\
    p_{a_{out}}&=p_{a_{in}}\cdot \frac{p_{\mathcal{A}_{out}}}{p_{\mathcal{A}_{in}}},\quad \forall a\in \mathcal{A}, \\
    p_{n_{out}}&= p_{n_{in}} \cdot (1 + \frac{p_{\mathcal{A}_{in}}-p_{\mathcal{A}_{out}}}{p_{\backslash \mathcal{A}_{in}}}),\quad \forall n\notin \mathcal{A},\\
\end{aligned}$}
\label{eq3}
\end{equation}

\noindent
where $p_{a_{in}}$ is the input probability of an attribute-related token $a$, $p_{a_{out}}$ is the output one, and the same goes for every non-attribute-related token $n \in \backslash \mathcal{A}$. Eq. \ref{eq3} first modifies the sum of the input probabilities of all attribute-related tokens $p_{\mathcal{A}_{in}}$, then rescales each attribute-related token $a$ by the ratio of $p_{\mathcal{A}_{out}}$ to $p_{\mathcal{A}_{in}}$, and finally rescales the probability of each non-attribute-related token $n$ by the ratio of the difference between $p_{\mathcal{A}_{in}}$ and $p_{\mathcal{A}_{out}}$ to the sum of the input probabilities of all non-attribute-related tokens $p_{\backslash \mathcal{A}_{in}}$. Fig. \ref{fig3} illustrates a simple example of language generation with \textsc{Gamma Sampling}.

\subsubsection{Controlling Multiple Attributes}

In situations where multiple attributes $\mathcal{A}^{1:T}=\{\mathcal{A}^{1},\mathcal{A}^{2},...,\mathcal{A}^{T}\}$ need to be controlled in a sequential manner, if simply utilizing the equation defined in Eq. \ref{eq3}, the earlier modifications may be overridden by later ones. To mitigate this issue, supplementary computations have been introduced to complement Eq. \ref{eq3}. When controlling the $t$-th attribute $\mathcal{A}^{t}$, the following processes are performed:

\begin{equation}
\resizebox{.85\hsize}{!}{$
\begin{aligned}
    \mathcal{F}^{t}\!&=\!\mathcal{A}^{1} \cup \mathcal{A}^{2} \cup ... \cup \mathcal{A}^{t-1} \!-\! \mathcal{A}^{t}, \\
    p_{\mathcal{A}^{t}_{out}}\!&=\!p_{\mathcal{A}^{t}_{in}}^{tan(\frac{\pi \Gamma^{t}}{2})}\cdot(1\!-\!p_{\mathcal{F}^{t}_{in}})^{1-tan(\frac{\pi \Gamma^{t}}{2})}, \\
    p_{a_{out}}\!&=\!p_{a_{in}}\cdot \frac{p_{\mathcal{A}^{t}_{out}}}{p_{\mathcal{A}^{t}_{in}}},\quad a\in \mathcal{A}^{t}, \\
    p_{n_{out}}\!&=\!p_{n_{in}} \cdot (1 \!+\!\frac{p_{\mathcal{A}^{t}_{in}}-p_{\mathcal{A}^{t}_{out}}}{p_{\backslash (\mathcal{A}^{t}\cup \mathcal{F}^{t})_{in}}}),\quad n\notin \mathcal{A}^{t}\cup\mathcal{F}^{t},\\
\end{aligned}$}
\label{eq4}
\end{equation}

\noindent
where $\Gamma^{t}$ represents the control strength for the attribute set $\mathcal{A}^{t}$, and the frozen set $\mathcal{F}^{t}$ consists of attribute-related tokens that have been modified in previous turns but not in $\mathcal{A}^{t}$. Like Eq. \ref{eq3}, the sum of token probabilities after modification in Eq. \ref{eq4} remains 1. The key distinction between the two equations is the inclusion of the frozen set $\mathcal{F}^{t}$, where token probabilities remain unchanged during the $t$-th turn.

\subsection{Controllable Attributes}

In our empirical research and analysis, we identified six controllable attributes at the token level. These attributes were utilized in Section 4 to explore their potential impact on generated text.

\textsc{Sentence Length} is an objective text attribute that is directly linked to ending marks such as full stops, question marks, and exclamation marks. By manipulating the frequency of these ending marks, the average sentence length can be effectively altered.

\textsc{Perfect Ending} is a dynamic tuning strategy of \textsc{Sentence Length}. When the number of generated words exceeds a certain threshold, the value of $\Gamma$ decreases linearly with the increase in the number of generated words. This process is designed to ensure that the text does not end abruptly.

\textsc{Topic Relevance} is controlled by selecting words that are most related to a user-given topic word, as calculated by the cosine similarity of word embedding in GPT2. In our specific implementation, we selected the top 100 tokens.

\textsc{Sentiment Polarity} is achieved by adjusting the frequency of words in two wordlists, each containing roughly 1000 words sourced from the internet. One list is comprised of negative words, while the other contains positive words.

\textsc{Repetition Decreasing}, similar to penalized sampling (\citealt{DBLP:journals/corr/abs-1909-05858}), prevents degeneration by decreasing the probabilities of tokens that have recently been generated.

\textsc{Relatedness Increasing} is a dynamic tuning strategy of \textsc{Topic Relevance} that aims to improve the coherence of text by encouraging the generation of words that are closely related to recently presented nouns labelled by NLTK (\citealt{DBLP:journals/corr/abs-1909-05858}).

By defining the attributes of interest at the token level, \textsc{Gamma Sampling} allows for effective and targeted manipulation of the generated text.

\section{Experiments}

To evaluate \textsc{Gamma Sampling}, a series of experiments were conducted. The metrics used in this evaluation are described in Section 4.1. In Section 4.2, we validate the fine-grained control achieved by \textsc{Gamma Sampling} and examine the impact of varying degrees of control strength. Finally, we conduct a comparative study in Section 4.3. The findings of this section better the understanding of \textsc{Gamma Sampling} and its potential applications in various domains.

\subsection{Metrics}
The evaluation of generated texts involves assessing four key aspects: fluency, diversity, attribute relevance, and overall quality. A comprehensive evaluation was conducted by utilizing both automatic metrics and human annotators sourced from Amazon's Mechanical Turk. To ensure the accuracy of the metrics, the text prompts, which include attribute information, were excluded from the experimental data, while the prefixes were retained.

\textbf{Fluency} \quad The fluency of the generated text is measured by GPT2 in different sizes, based on perplexity (\citealt{DBLP:journals/corr/VinyalsL15}). A higher \textbf{PPL}, indicates that it is less likely that GPT2 will generate such text. It is worth noting that a lower PPL does not necessarily imply better quality, as degenerated text usually has a very low PPL, which is deemed to be of poor quality.

\textbf{Diversity} \quad The diversity of the generated samples is evaluated using \textbf{DIST-N} scores (\citealt{DBLP:conf/naacl/LiGBGD16}). A higher value of DIST-N indicates a higher proportion of distinct N-grams.

\textbf{Attribute Relevance} \quad We focused on evaluating three types of controllable generation: \textsc{Sentence Length}, \textsc{Topic Relevance} and \textsc{Sentiment Polarity}. The average sentence length (\textbf{ASL}) measures the average number of words per sentence. The latter two are evaluated using two metrics based on external classifiers\footnote{\url{https://huggingface.co/facebook/bart-large-mnli}}\footnote{\url{https://huggingface.co/cardiffnlp/twitter-roberta-base-sentiment-latest}}: external classifier accuracy (\textbf{ECA}) and external classifier confidence (\textbf{ECC}). The higher the ECA and ECC, the more salient the external classifier considers the generated text to hold a certain topic/sentiment. 

\textbf{Overall Quality} \quad  The overall quality of machine-generated text was evaluated based on the TOEFL Independent Writing Rubrics\footnote{\url{https://www.ets.org/content/dam/ets-org/pdfs/toefl/toefl-ibt-writing-rubrics.pdf}}, which are widely used to assess the English writing skills of non-native speakers. However, additional considerations were taken into account to ensure a comprehensive evaluation of the machine-generated text. Specifically, we do not take into account whether the text itself is finished or contains factual errors, but extensive repetition, completely off-topic, or obvious common sense errors would result in a low score. This revised set of criteria is TOEFL Writing Rubrics for Machine-generated Text (\textbf{T4MT}). MTurkers are asked to rate each generated text on a scale of 0 to 5 (nonsense to advanced) based on T4MT. More information about this scoring criterion can be found in \nameref{Appendix A}.

\begin{figure}[t]
    \hspace{-1em}
	\centering
		\begin{minipage}{8cm} 
            \includegraphics[width=1\linewidth]{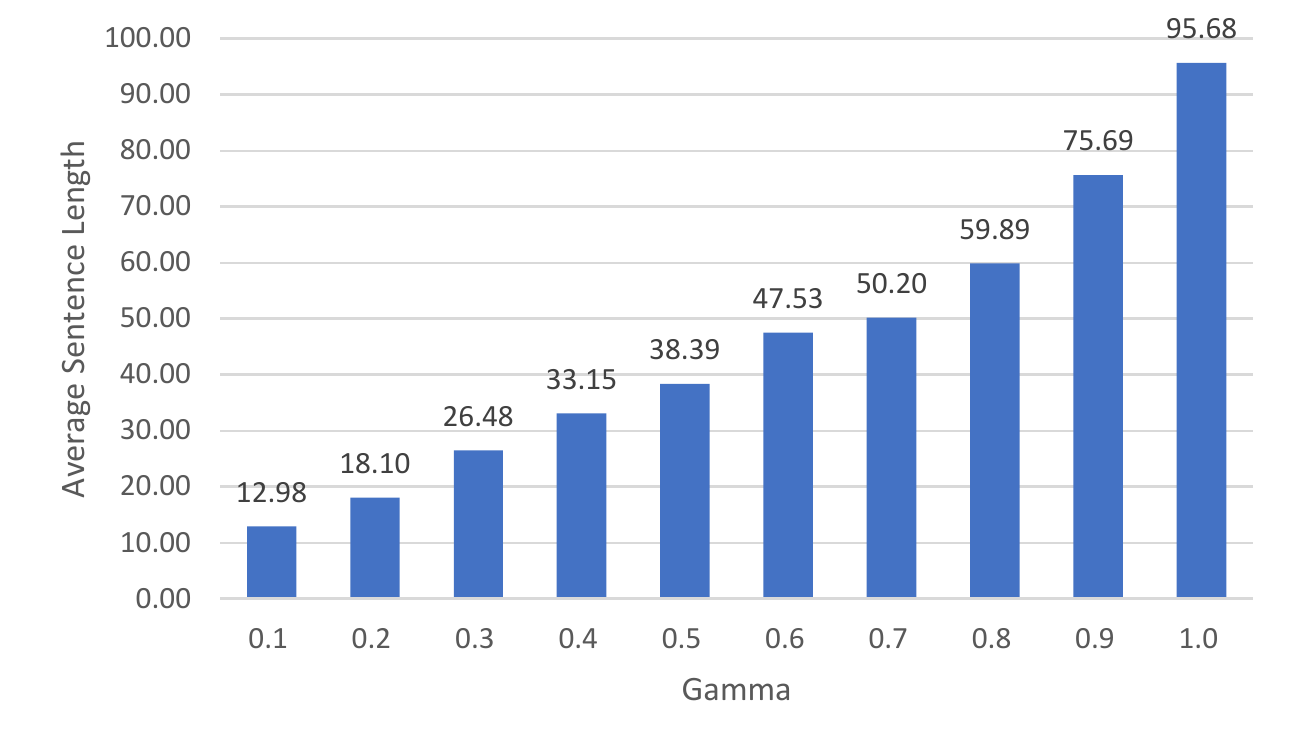}
        \end{minipage}
    \centering
    \vspace{-1em}
	\caption{Average Sentence Length (ASL) in various settings of $\Gamma$ (Gamma).}
	\label{fig5}
    \vspace{-1em}
\end{figure}

\subsection{Control and Over-Control Analysis}
In these experiments, we evaluated GPT2-Small on \textsc{Sentence Length}. The generation process was initiated with the prefix "The issue focused" and the model was instructed to generate the subsequent 100 tokens. To guarantee that each sample contained at least one ending mark, we implemented \textsc{Perfect Ending}. This involved linearly decreasing the value of $\Gamma$ to 0 as the number of generated tokens increased, starting from the 80th token. Representative samples of \textsc{Sentence Length} can be found in \nameref{Appendix B}.

We first investigated fine-grained control using \textsc{Gamma Sampling}. Specifically, 100 samples were generated from GPT2 for each of ten different values of $\Gamma$, resulting in a total of 1000 samples. Our results show a clear correlation between $\Gamma$ and the ASL of the generated text. Fig. \ref{fig5} illustrates that ASL increases significantly with increasing $\Gamma$, from an average of 12.98 words/sentence at $\Gamma=0.1$ to an average of 95.68 words/sentence at $\Gamma=1.0$. For comparison, the ASL of text generated by GPT2-Small is 38.39 at $\Gamma=0.5$. This result reveals a linear relationship between ASL and $\Gamma$, suggesting that the probability scaling of attribute-related tokens has a non-linear impact on the corresponding attribute. Additionally, without \textsc{Perfect Ending}, sentences will not end when $\Gamma=1.0$.

\begin{table}[t!]
  \begin{center}
    \caption{Results of \textsc{Sentence Length} in various settings of $\Gamma$ and top-$p$.}

    \resizebox{\linewidth}{!}{\begin{tabular}{l|c|c|c|c|c}
      \toprule % <-- Toprule here
      \textbf{top-$p$, $\Gamma$} & \textbf{PPL-Small↓} & \textbf{DIST-1↑} & \textbf{DIST-2↑} & \textbf{DIST-3↑} & \textbf{ASL}\\
      \midrule % <-- Midrule here
      \textbf{1.0, 0.9} & 64.65 & 62.90 & 78.02 & 80.71 & 75.69\\
      \textbf{0.9, 0.9} & 21.33 & 59.72 & 76.85 & 80.68 & 59.21\\
      \textbf{0.8, 0.9} & 12.49 & 55.24 & 72.32 & 76.18 & 53.92\\
      \textbf{1.0, 0.5} & 56.95 & 62.27 & 78.22 & 81.48 & 38.39\\
      \textbf{0.9, 0.5} & 20.42 & 60.98 & 78.34 & 81.94 & 31.54\\
      \textbf{0.8, 0.5} & 15.14 & 58.11 & 76.20 & 80.56 & 25.33\\
      \textbf{1.0, 0.1} & 6844.10 & 64.40 & 69.60 & 68.37 & 12.98\\
      \textbf{0.9, 0.1} & 48.75 & 51.59 & 61.98 & 64.19 & 15.25\\
      \textbf{0.8, 0.1} & 16.53 & 48.08 & 61.00 & 64.10 & 10.84\\
      \bottomrule % <-- Bottomrule here
    \end{tabular}}
  \end{center}
  \vspace{-1em}
\end{table}

We then assessed the risk of over-control strength. To achieve this, we conducted an experiment wherein we selected three representative values of $\Gamma$ (0.1, 0.5, and 0.9) and applied nucleus sampling with varying top-$p$ (0.8, 0.9 and 1.0). We generated 100 samples in each of the nine settings, and as we hypothesized, we found that increasing the probability of controlled tokens led to a significant decrease in the quality of the generated text, as evidenced by Table 1. With a top-$p$ value of 1.0 and a $\Gamma$ value of 0.1, the PPL of the generated samples reached an astonishingly high value of 6844.1, indicating that it would be extremely difficult for GPT2-Small to generate such text. However, we observed that reducing the top-$p$ value effectively mitigated the negative impact of \textsc{Gamma Sampling} on the quality of generated text. For example, with a top-$p$ value of 0.8 and a $\Gamma$ value of 0.1, the PPL dropped dramatically to 16.53. In contrast, the impact of drastically reducing the probability of controlled tokens ($\Gamma=0.9$) was much less significant.

In conclusion, the experimental results demonstrate that \textsc{Gamma Sampling} can effectively modify the attributes of generated text. To mitigate the negative effects of over-control strength, pre-sampling should be applied.

\begin{table*}[t!]
    \vspace{1em}
  \begin{center}
    \caption{The main results for comparing all the methods for controllable language generation. The best results for each metric are highlighted in bold and red for clarity. In the interest of space, we present the results for GPT2, GS, and GSM as averages across all sizes of GPT2. Additionally, PPL is also presented as an average across all sizes of GPT2. For more detailed results, we refer readers to \nameref{Appendix D}.}

    \begin{tabular}{l|c|c|c|c|c|c|c}
      \toprule % <-- Toprule here
      \textbf{Method} & \textbf{PPL↓} & \textbf{DIST-1↑} & \textbf{DIST-2↑} & \textbf{DIST-3↑} & \textbf{ECA↑} & \textbf{ECC↑} & \textbf{T4MT↑}\\
      \midrule % <-- Midrule here
      \textbf{GPT2} & 11.71 & 44.85 & 59.49 & 64.19 & 58.56 & 51.39 & 2.71±0.96\\
      \textbf{GS} & \textbf{\textcolor{red}{8.24}} & 36.52 & 49.81 & 54.60 & 56.79 & 51.46 & 2.76±1.01\\
      \textbf{GSM} & 23.74 & \textbf{\textcolor{red}{78.10}} & \textbf{\textcolor{red}{93.72}} & 95.17 & \textbf{\textcolor{red}{60.73}} & \textbf{\textcolor{red}{54.03}} & \textbf{\textcolor{red}{3.35±0.77}}\\
      \textbf{GPT2-FT} & 27.61 & 53.90 & 71.96 & 76.88 & 48.59 & 44.53 & 3.08±0.93\\
      \textbf{PPLM-BCR} & 11.01 & 60.79 & 85.80 & 90.54 & 43.50 & 40.08 & 3.18±0.83\\
      \textbf{K2T} & 20.24 & 69.93 & 93.59 & \textbf{\textcolor{red}{96.32}} & 39.50 & 38.08 & 3.15±0.80\\
      \textbf{CTRL} & 13.37 & 46.74 & 65.74 & 71.54 & 51.67 & 45.38 & 3.12±0.89\\
      \bottomrule % <-- Bottomrule here
    \end{tabular}
  \end{center}
\end{table*}

\begin{figure*}[t]
	Table 3: Comparison of different methods for controllable language generation. PPLM, K2T, and \textsc{Gamma Sampling} are methods that do not require the use of prompts as they do not introduce attribute-related information at the input. It is important to note that the generation time of K2T increases linearly with the number of keywords used. In contrast, the time cost of \textsc{Gamma Sampling} is insensitive to the number of controlled tokens, resulting in similar generation times for GS and GSM.
	
	\centering
        \includegraphics[width=1\linewidth]{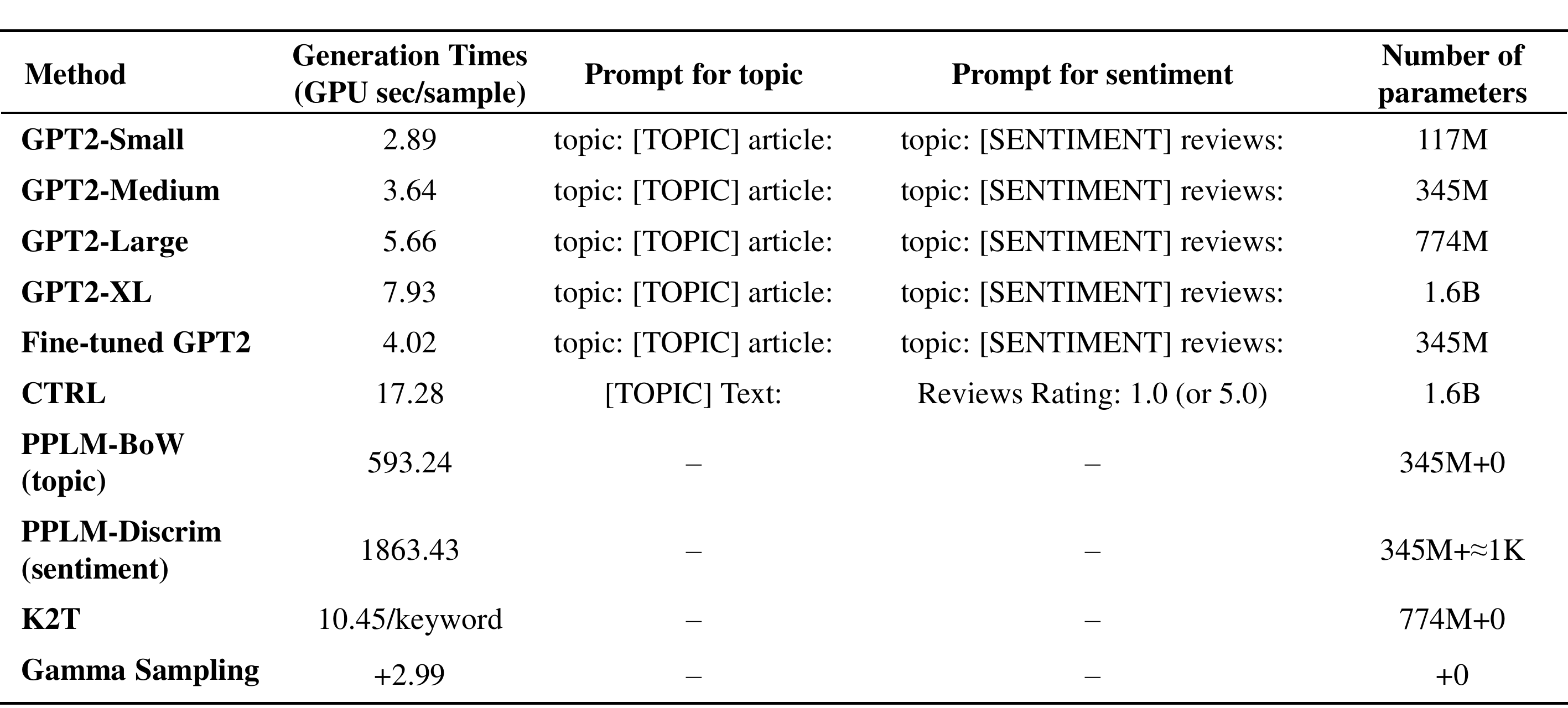}
    \centering
    \vspace{-2em}
\end{figure*}

\subsection{Comparative Study}

As shown in Table 2, the methods evaluated based on \textsc{Topic Relevance} and \textsc{Sentiment Polarity} in this study include:

\textbf{GPT2} (\citealt{radford2019language}) \quad Vanilla GPT2 models in four sizes (Small, Medium, Large, and XL) were used, with attribute-related information being introduced via prompts.

\textbf{GS} \quad A baseline of \textsc{Gamma Sampling} that only enables \textsc{Topic Relevance} or \textsc{Sentiment Polarity} to control GPT2 in four sizes, with the probability of attribute-related tokens increased by $\Gamma=0.1$.

\textbf{GSM} \quad In addition to \textsc{Topic Relevance} or \textsc{Sentiment Polarity}, this method enables multiple controllable attributes, i.e., \textsc{Repetition Decreasing} ($\Gamma=0.9$) and \textsc{Relatedness Increasing} ($\Gamma=0.3$).

\textbf{GPT2-FT}\footnote{\url{https://huggingface.co/ktrapeznikov/gpt2-medium-topic-news}} \quad GPT2-Medium fine-tuned on a large news corpus for topic-controllable generation. This method demonstrates good zero-shot generation performance as long as the topic is a single word and not too specific.

\textbf{PPLM-BCR} (\citealt{DBLP:conf/iclr/DathathriMLHFMY20}) \quad This method guides GPT2-Medium and has two different schemes for \textsc{Topic Relevance} and \textsc{Sentiment Polarity}. Both of these schemes produce 10 samples, and the optimal sample is chosen based on its log-likelihood. The combined reference to both schemes is denoted as PPLM-BCR.

\textbf{K2T} (\citealt{DBLP:conf/emnlp/PascualEMCW21}) \quad A guided decoding method modifies the score function to encourage GPT2-Large to generate more words that are semantically similar to a given keyword set. The selected keyword set was the same as that used for \textsc{Gamma Sampling} to ensure comparability. However, as the number of controlled tokens increases, the time cost of this method rises linearly, and the generation time becomes unacceptable when all keywords from the \textsc{Sentiment Polarity} ($\approx$10,000 sec/sample) are used. Following the default setting for this method, a subset of 5 attribute-related tokens was randomly selected as keywords for each K2T-generated sample.

\textbf{CTRL} (\citealt{DBLP:journals/corr/abs-1909-05858}) \quad A conditional transformer-based LM trained from scratch to condition control codes, also known as prompts. This model has 1.6 billion parameters.

In this experiment, we followed the experimental procedures established by PPLM (\citealt{DBLP:conf/iclr/DathathriMLHFMY20}) and investigated the performance of models across three topics, namely \texttt{COMPUTERS}, \texttt{LEGAL}, and \texttt{SCIENCE}, as well as two sentiment polarities, \texttt{NEGATIVE} and \texttt{POSITIVE}. Given ethical considerations, we excluded topics related to \texttt{POLITICS} and \texttt{RELIGION}. For each task, 100 essays were generated by each model, with a length of 100 words per essay. To generate the essays, we used prefixes such as \texttt{The issue focused} for topics and \texttt{The movie} for sentiments, along with their respective prompts if available. For all methods in this experiment, we applied nucleus sampling with a top-$p=0.8$. In total, 8000 samples were generated for automatic evaluation, and 2400 samples were selected for human evaluation. Specifically, we selected the first 30 generations for each model per task, resulting in a total of 7285 valid ratings, with each sample being reviewed by at least 3 MTurkers. For further details, see \nameref{Appendix C} for samples from all methods.

The results presented in Table 2 indicate that the PPL of GS-generated text is lower than that of GPT2, which suggests that the former produces less complex text. However, the GS-generated text displays the lowest diversity as measured by DIST-N. Upon manual examination, we found that GS-generated text tended to suffer from a higher degree of degeneration compared to GPT2-generated text, with a higher likelihood of repeating words with similar semantics to the topic words. It indicates that using only \textsc{Topic Relevance} or \textsc{Sentiment Polarity} to control GPT2 did not yield favourable results in the controllable language generation task. In contrast, we found that GSM produced highly diverse text as measured by DIST-N, while maintaining a high attribute relevance as indicated by the ECA and ECC. Although there was a slight decrease in fluency as measured by PPL, the quality of the text generated by GSM remained acceptable as suggested by T4MT. These findings suggest that using both \textsc{Repetition Decreasing} and \textsc{Relatedness Increasing} together can significantly improve the quality of the generated text without going off-distribution.

In terms of T4MT, most methods demonstrated comparable results, while independent samples $t$-tests revealed that the GSM results were statistically significant when compared to all other methods, with a $p$-value $<$ 0.05 (excluding PPLM-BCR, which had a $p$-value of 0.051). Although we cannot assert the statistical superiority of GSM-generated text over that of PPLM-BCR, it is notable that the generation time of PPLM-BCR is at least 100 $\times$ longer than that of GSM, as indicated in Table 3.

The guided decoding baseline, K2T, has been found to generate text that is more off-topic when compared to \textsc{Gamma Sampling} as measured by ECA and ECC. This difference in performance is likely because K2T used smaller keyword sets (only 5 words/set) and a weaker control strength to prevent off-distribution issues. In contrast, the generation time of \textsc{Gamma Sampling} remains relatively consistent, and the negative impacts of strong control strength are effectively mitigated through the use of pre-sampling.

\section{Conclusions}
In this paper, we introduce a new technique called \textsc{Gamma Sampling}, which provides a highly efficient method for fine-grained control of LMs at the token level. The technique involves incorporating attribute-related information at the decoding time and enables the pre-definition or selection of attributes by the LMs themselves without requiring any additional training data. Experimental results demonstrate that \textsc{Gamma Sampling} is highly effective in controlling GPT2, producing improved objective and subjective metrics while maintaining a fast generation speed. Additionally, our method has been shown to be computationally efficient and able to avoid off-distribution issues. Since \textsc{Gamma Sampling} is a training-free approach, it is easily adaptable to any LM and has the potential to enhance the efficiency and user-friendliness of controllable language generation.

\section*{Limitations}

While \textsc{Gamma Sampling} has been shown to be an effective method for fine-grained controlling of LMs,  there is still room for improvement and further research in this area.

As a guided decoding method, \textsc{Gamma Sampling} relies on pre-defined or selected attribute-related tokens to guide the generation process. This means that the quality of the generated text is dependent on the quality of the selection of tokens. Furthermore, the attribute-related tokens must be selected by humans or LMs themselves, which may be prone to errors or bias. Improving the reliability and robustness of attribute information is an important future direction.

In addition, \textsc{Gamma Sampling} is currently only able to control attributes at the token level. While this allows for fine-grained control, it may be limited in its ability to control more complex or abstract attributes (e.g., writing style). Developing methods to control these types of attributes is a key area for future research.

\section*{Ethics Statement}

As with any technology, there are potential ethical considerations to be taken into account with the use of \textsc{Gamma Sampling}. One of the issues we have identified in conducting experiments is the potential for bias in the attribute information provided to guide the generation process. If the attribute information is provided by humans, it may be prone to bias based on the human annotator's own beliefs, experiences, or stereotypes. On the other hand, if the attribute information is selected by the LMs themselves, it may be like an echo chamber, which further amplifies any biases present in the training data used to train the LMs. For this reason, we excluded topics where models could generate offensive text (i.e., \texttt{POLITICS} and \texttt{RELIGION}).

To mitigate this risk, we recommend that attribute-related tokens be carefully curated and checked for bias before being used in the generation process. In addition, it is important to ensure that the selection of attribute-related tokens is diverse and inclusive, to avoid perpetuating stereotypes or discrimination.

\bibliography{anthology,custom}

\begin{thebibliography}{29}
\expandafter\ifx\csname natexlab\endcsname\relax\def\natexlab#1{#1}\fi

\bibitem[{Applebaum(1952)}]{applebaum1952gamma}
Sidney Applebaum. 1952.
\newblock Gamma correction in constant luminance color television systems.
\newblock \emph{Proceedings of the IRE}, 40(10):1185--1195.

\bibitem[{Dabre and Fujita(2021)}]{DBLP:conf/mtsummit/DabreF21}
Raj Dabre and Atsushi Fujita. 2021.
\newblock \href {https://aclanthology.org/2021.mtsummit-research.10}
  {Investigating softmax tempering for training neural machine translation
  models}.
\newblock In \emph{Proceedings of the 18th Biennial Machine Translation Summit
  - Volume 1: Research Track, MTSummit 2021 Virtual, August 16-20, 2021}, pages
  114--126. Association for Machine Translation in the Americas.

\bibitem[{Dathathri et~al.(2020)Dathathri, Madotto, Lan, Hung, Frank, Molino,
  Yosinski, and Liu}]{DBLP:conf/iclr/DathathriMLHFMY20}
Sumanth Dathathri, Andrea Madotto, Janice Lan, Jane Hung, Eric Frank, Piero
  Molino, Jason Yosinski, and Rosanne Liu. 2020.
\newblock \href {https://openreview.net/forum?id=H1edEyBKDS} {Plug and play
  language models: {A} simple approach to controlled text generation}.
\newblock In \emph{8th International Conference on Learning Representations,
  {ICLR} 2020, Addis Ababa, Ethiopia, April 26-30, 2020}. OpenReview.net.

\bibitem[{Fan et~al.(2018)Fan, Lewis, and Dauphin}]{DBLP:conf/acl/LewisDF18}
Angela Fan, Mike Lewis, and Yann~N. Dauphin. 2018.
\newblock \href {https://doi.org/10.18653/v1/P18-1082} {Hierarchical neural
  story generation}.
\newblock In \emph{Proceedings of the 56th Annual Meeting of the Association
  for Computational Linguistics, {ACL} 2018, Melbourne, Australia, July 15-20,
  2018, Volume 1: Long Papers}, pages 889--898. Association for Computational
  Linguistics.

\bibitem[{Gehman et~al.(2020)Gehman, Gururangan, Sap, Choi, and
  Smith}]{DBLP:conf/emnlp/GehmanGSCS20}
Samuel Gehman, Suchin Gururangan, Maarten Sap, Yejin Choi, and Noah~A. Smith.
  2020.
\newblock \href {https://doi.org/10.18653/v1/2020.findings-emnlp.301}
  {Realtoxicityprompts: Evaluating neural toxic degeneration in language
  models}.
\newblock In \emph{Findings of the Association for Computational Linguistics:
  {EMNLP} 2020, Online Event, 16-20 November 2020}, volume {EMNLP} 2020 of
  \emph{Findings of {ACL}}, pages 3356--3369. Association for Computational
  Linguistics.

\bibitem[{Ghazvininejad et~al.(2017)Ghazvininejad, Shi, Priyadarshi, and
  Knight}]{DBLP:conf/acl/GhazvininejadSP17}
Marjan Ghazvininejad, Xing Shi, Jay Priyadarshi, and Kevin Knight. 2017.
\newblock \href {https://doi.org/10.18653/v1/P17-4008} {Hafez: an interactive
  poetry generation system}.
\newblock In \emph{Proceedings of the 55th Annual Meeting of the Association
  for Computational Linguistics, {ACL} 2017, Vancouver, Canada, July 30 -
  August 4, System Demonstrations}, pages 43--48. Association for Computational
  Linguistics.

\bibitem[{Ghosh et~al.(2017)Ghosh, Chollet, Laksana, Morency, and
  Scherer}]{DBLP:conf/acl/GhoshCLMS17}
Sayan Ghosh, Mathieu Chollet, Eugene Laksana, Louis{-}Philippe Morency, and
  Stefan Scherer. 2017.
\newblock \href {https://doi.org/10.18653/v1/P17-1059} {Affect-lm: {A} neural
  language model for customizable affective text generation}.
\newblock In \emph{Proceedings of the 55th Annual Meeting of the Association
  for Computational Linguistics, {ACL} 2017, Vancouver, Canada, July 30 -
  August 4, Volume 1: Long Papers}, pages 634--642. Association for
  Computational Linguistics.

\bibitem[{Holtzman et~al.(2020)Holtzman, Buys, Du, Forbes, and
  Choi}]{DBLP:conf/iclr/HoltzmanBDFC20}
Ari Holtzman, Jan Buys, Li~Du, Maxwell Forbes, and Yejin Choi. 2020.
\newblock \href {https://openreview.net/forum?id=rygGQyrFvH} {The curious case
  of neural text degeneration}.
\newblock In \emph{8th International Conference on Learning Representations,
  {ICLR} 2020, Addis Ababa, Ethiopia, April 26-30, 2020}. OpenReview.net.

\bibitem[{Keskar et~al.(2019)Keskar, McCann, Varshney, Xiong, and
  Socher}]{DBLP:journals/corr/abs-1909-05858}
Nitish~Shirish Keskar, Bryan McCann, Lav~R. Varshney, Caiming Xiong, and
  Richard Socher. 2019.
\newblock \href {http://arxiv.org/abs/1909.05858} {{CTRL:} {A} conditional
  transformer language model for controllable generation}.
\newblock \emph{CoRR}, abs/1909.05858.

\bibitem[{Kojima et~al.(2022)Kojima, Gu, Reid, Matsuo, and
  Iwasawa}]{DBLP:journals/corr/abs-2205-11916}
Takeshi Kojima, Shixiang~Shane Gu, Machel Reid, Yutaka Matsuo, and Yusuke
  Iwasawa. 2022.
\newblock \href {https://doi.org/10.48550/arXiv.2205.11916} {Large language
  models are zero-shot reasoners}.
\newblock \emph{CoRR}, abs/2205.11916.

\bibitem[{Krause et~al.(2021)Krause, Gotmare, McCann, Keskar, Joty, Socher, and
  Rajani}]{DBLP:conf/emnlp/KrauseGMKJSR21}
Ben Krause, Akhilesh~Deepak Gotmare, Bryan McCann, Nitish~Shirish Keskar,
  Shafiq~R. Joty, Richard Socher, and Nazneen~Fatema Rajani. 2021.
\newblock \href {https://doi.org/10.18653/v1/2021.findings-emnlp.424} {Gedi:
  Generative discriminator guided sequence generation}.
\newblock In \emph{Findings of the Association for Computational Linguistics:
  {EMNLP} 2021, Virtual Event / Punta Cana, Dominican Republic, 16-20 November,
  2021}, pages 4929--4952. Association for Computational Linguistics.

\bibitem[{Li et~al.(2016)Li, Galley, Brockett, Gao, and
  Dolan}]{DBLP:conf/naacl/LiGBGD16}
Jiwei Li, Michel Galley, Chris Brockett, Jianfeng Gao, and Bill Dolan. 2016.
\newblock \href {https://doi.org/10.18653/v1/n16-1014} {A diversity-promoting
  objective function for neural conversation models}.
\newblock In \emph{{NAACL} {HLT} 2016, The 2016 Conference of the North
  American Chapter of the Association for Computational Linguistics: Human
  Language Technologies, San Diego California, USA, June 12-17, 2016}, pages
  110--119. The Association for Computational Linguistics.

\bibitem[{Liu et~al.(2021)Liu, Sap, Lu, Swayamdipta, Bhagavatula, Smith, and
  Choi}]{DBLP:conf/acl/LiuSLSBSC20}
Alisa Liu, Maarten Sap, Ximing Lu, Swabha Swayamdipta, Chandra Bhagavatula,
  Noah~A. Smith, and Yejin Choi. 2021.
\newblock \href {https://doi.org/10.18653/v1/2021.acl-long.522} {Dexperts:
  Decoding-time controlled text generation with experts and anti-experts}.
\newblock In \emph{Proceedings of the 59th Annual Meeting of the Association
  for Computational Linguistics and the 11th International Joint Conference on
  Natural Language Processing, {ACL/IJCNLP} 2021, (Volume 1: Long Papers),
  Virtual Event, August 1-6, 2021}, pages 6691--6706. Association for
  Computational Linguistics.

\bibitem[{Lu et~al.(2021)Lu, Welleck, West, Jiang, Kasai, Khashabi, Bras, Qin,
  Yu, Zellers, Smith, and Choi}]{DBLP:journals/corr/abs-2112-08726}
Ximing Lu, Sean Welleck, Peter West, Liwei Jiang, Jungo Kasai, Daniel Khashabi,
  Ronan~Le Bras, Lianhui Qin, Youngjae Yu, Rowan Zellers, Noah~A. Smith, and
  Yejin Choi. 2021.
\newblock \href {http://arxiv.org/abs/2112.08726} {Neurologic a*esque decoding:
  Constrained text generation with lookahead heuristics}.
\newblock \emph{CoRR}, abs/2112.08726.

\bibitem[{Ma et~al.(2020)Ma, Sap, Rashkin, and Choi}]{DBLP:conf/emnlp/MaSRC20}
Xinyao Ma, Maarten Sap, Hannah Rashkin, and Yejin Choi. 2020.
\newblock \href {https://doi.org/10.18653/v1/2020.emnlp-main.602}
  {Powertransformer: Unsupervised controllable revision for biased language
  correction}.
\newblock In \emph{Proceedings of the 2020 Conference on Empirical Methods in
  Natural Language Processing, {EMNLP} 2020, Online, November 16-20, 2020},
  pages 7426--7441. Association for Computational Linguistics.

\bibitem[{Meng et~al.(2022)Meng, Lu, Peng, and
  Chang}]{DBLP:journals/corr/abs-2205-14219}
Tao Meng, Sidi Lu, Nanyun Peng, and Kai{-}Wei Chang. 2022.
\newblock \href {https://doi.org/10.48550/arXiv.2205.14219} {Controllable text
  generation with neurally-decomposed oracle}.
\newblock \emph{CoRR}, abs/2205.14219.

\bibitem[{Mireshghallah et~al.(2022)Mireshghallah, Goyal, and
  Berg{-}Kirkpatrick}]{DBLP:journals/corr/abs-2203-13299}
Fatemehsadat Mireshghallah, Kartik Goyal, and Taylor Berg{-}Kirkpatrick. 2022.
\newblock \href {https://doi.org/10.48550/arXiv.2203.13299} {Mix and match:
  Learning-free controllable text generation using energy language models}.
\newblock \emph{CoRR}, abs/2203.13299.

\bibitem[{Neri(2009)}]{neri2009nonlinear}
Peter Neri. 2009.
\newblock Nonlinear characterization of a simple process in human vision.
\newblock \emph{Journal of Vision}, 9(12):1--1.

\bibitem[{Pascual et~al.(2020)Pascual, Egressy, Bolli, and
  Wattenhofer}]{DBLP:journals/corr/abs-2012-15416}
Damian Pascual, Beni Egressy, Florian Bolli, and Roger Wattenhofer. 2020.
\newblock \href {http://arxiv.org/abs/2012.15416} {Directed beam search:
  Plug-and-play lexically constrained language generation}.
\newblock \emph{CoRR}, abs/2012.15416.

\bibitem[{Pascual et~al.(2021)Pascual, Egressy, Meister, Cotterell, and
  Wattenhofer}]{DBLP:conf/emnlp/PascualEMCW21}
Damian Pascual, Beni Egressy, Clara Meister, Ryan Cotterell, and Roger
  Wattenhofer. 2021.
\newblock \href {https://doi.org/10.18653/v1/2021.findings-emnlp.334} {A
  plug-and-play method for controlled text generation}.
\newblock In \emph{Findings of the Association for Computational Linguistics:
  {EMNLP} 2021, Virtual Event / Punta Cana, Dominican Republic, 16-20 November,
  2021}, pages 3973--3997. Association for Computational Linguistics.

\bibitem[{Powers(1998)}]{DBLP:conf/conll/Powers98}
David M.~W. Powers. 1998.
\newblock \href {https://aclanthology.org/W98-1218/} {Applications and
  explanations of zipf's law}.
\newblock In \emph{Proceedings of the Joint Conference on New Methods in
  Language Processing and Computational Natural Language Learning, NeMLaP/CoNLL
  1998, Macquarie University, Sydney, NSW, Australia, January 11-17, 1998},
  pages 151--160. {ACL}.

\bibitem[{Radford et~al.(2019)Radford, Wu, Child, Luan, Amodei, Sutskever
  et~al.}]{radford2019language}
Alec Radford, Jeffrey Wu, Rewon Child, David Luan, Dario Amodei, Ilya
  Sutskever, et~al. 2019.
\newblock Language models are unsupervised multitask learners.
\newblock \emph{OpenAI blog}, 1(8):9.

\bibitem[{Shin et~al.(2020)Shin, Razeghi, IV, Wallace, and
  Singh}]{DBLP:conf/emnlp/ShinRLWS20}
Taylor Shin, Yasaman Razeghi, Robert L.~Logan IV, Eric Wallace, and Sameer
  Singh. 2020.
\newblock \href {https://doi.org/10.18653/v1/2020.emnlp-main.346} {Autoprompt:
  Eliciting knowledge from language models with automatically generated
  prompts}.
\newblock In \emph{Proceedings of the 2020 Conference on Empirical Methods in
  Natural Language Processing, {EMNLP} 2020, Online, November 16-20, 2020},
  pages 4222--4235. Association for Computational Linguistics.

\bibitem[{Vinyals and Le(2015)}]{DBLP:journals/corr/VinyalsL15}
Oriol Vinyals and Quoc~V. Le. 2015.
\newblock \href {http://arxiv.org/abs/1506.05869} {A neural conversational
  model}.
\newblock \emph{CoRR}, abs/1506.05869.

\bibitem[{Weng(2021)}]{weng2021conditional}
Lilian Weng. 2021.
\newblock \href
  {https://lilianweng.github.io/posts/2021-01-02-controllable-text-generation/}
  {Controllable neural text generation.}
\newblock \emph{lilianweng.github.io}.

\bibitem[{Xu et~al.(2021)Xu, Luo, Zhang, Tan, Chang, Huang, and
  Huang}]{DBLP:conf/emnlp/XuLZTCHH21}
Runxin Xu, Fuli Luo, Zhiyuan Zhang, Chuanqi Tan, Baobao Chang, Songfang Huang,
  and Fei Huang. 2021.
\newblock \href {https://doi.org/10.18653/v1/2021.emnlp-main.749} {Raise a
  child in large language model: Towards effective and generalizable
  fine-tuning}.
\newblock In \emph{Proceedings of the 2021 Conference on Empirical Methods in
  Natural Language Processing, {EMNLP} 2021, Virtual Event / Punta Cana,
  Dominican Republic, 7-11 November, 2021}, pages 9514--9528. Association for
  Computational Linguistics.

\bibitem[{Yang and Klein(2021)}]{DBLP:conf/naacl/YangK21}
Kevin Yang and Dan Klein. 2021.
\newblock \href {https://doi.org/10.18653/v1/2021.naacl-main.276} {{FUDGE:}
  controlled text generation with future discriminators}.
\newblock In \emph{Proceedings of the 2021 Conference of the North American
  Chapter of the Association for Computational Linguistics: Human Language
  Technologies, {NAACL-HLT} 2021, Online, June 6-11, 2021}, pages 3511--3535.
  Association for Computational Linguistics.

\bibitem[{Ziegler et~al.(2019)Ziegler, Stiennon, Wu, Brown, Radford, Amodei,
  Christiano, and Irving}]{DBLP:journals/corr/abs-1909-08593}
Daniel~M. Ziegler, Nisan Stiennon, Jeffrey Wu, Tom~B. Brown, Alec Radford,
  Dario Amodei, Paul~F. Christiano, and Geoffrey Irving. 2019.
\newblock \href {http://arxiv.org/abs/1909.08593} {Fine-tuning language models
  from human preferences}.
\newblock \emph{CoRR}, abs/1909.08593.

\bibitem[{Zou et~al.(2021)Zou, Yin, Zhong, Yang, Yang, and
  Tang}]{DBLP:conf/kdd/ZouYZYYT21}
Xu~Zou, Da~Yin, Qingyang Zhong, Hongxia Yang, Zhilin Yang, and Jie Tang. 2021.
\newblock \href {https://doi.org/10.1145/3447548.3467418} {Controllable
  generation from pre-trained language models via inverse prompting}.
\newblock In \emph{{KDD} '21: The 27th {ACM} {SIGKDD} Conference on Knowledge
  Discovery and Data Mining, Virtual Event, Singapore, August 14-18, 2021},
  pages 2450--2460. {ACM}.

\end{thebibliography}
\bibliographystyle{acl_natbib}

\vfill
\onecolumn
\newpage
\appendix
\section*{Appendix A}\label{Appendix A}
\textbf{\small Please read the “General Rules” below carefully to ensure that you understand the scoring criteria. More details are given in the “Writing Rubrics” (in full instructions). If your score is too far away from the average of other annotators (difference in absolute value greater than 1), it will be rejected.}
  \\
  \\
\textbf{\small Notice:} \small The contents of these essays are not necessarily factual so do not believe exactly what you read. If an essay contains a link to a web page, do not access it in your browser.
  \\
  \\
\textbf{General Rules:}

\begin{itemize}
    \item 
    \small Each essay needs to be rated on a scale of 0 to 5 depending on its quality and whether they fit the given topic
    
    \item
    \small Most of the essays are around 100 words in length, and most of them stop mid-sentence (essays are not finished). It is common and does not affect the scoring
    
    \item
    \small If an essay is completely off-topic, or included something that clearly contradicts common sense, only a maximum of 1 mark can be scored
    
    \item
    \small If an essay contains factual errors (e.g., South African anti-Apartheid leader Nelson Mandela dying in prison in the 1980s) but not common sense errors (e.g., the sun rises in the west), it does not affect the scoring
    
\end{itemize}

\textbf{Writing Rubrics\footnote{mainly based on TOEFL iBT Independent Writing Rubrics\\
\url{https://www.ets.org/content/dam/ets-org/pdfs/toefl/toefl-ibt-writing-rubrics.pdf}}:}
\begin{figure}[h]
	\centering
        \includegraphics[width=1.01\linewidth]{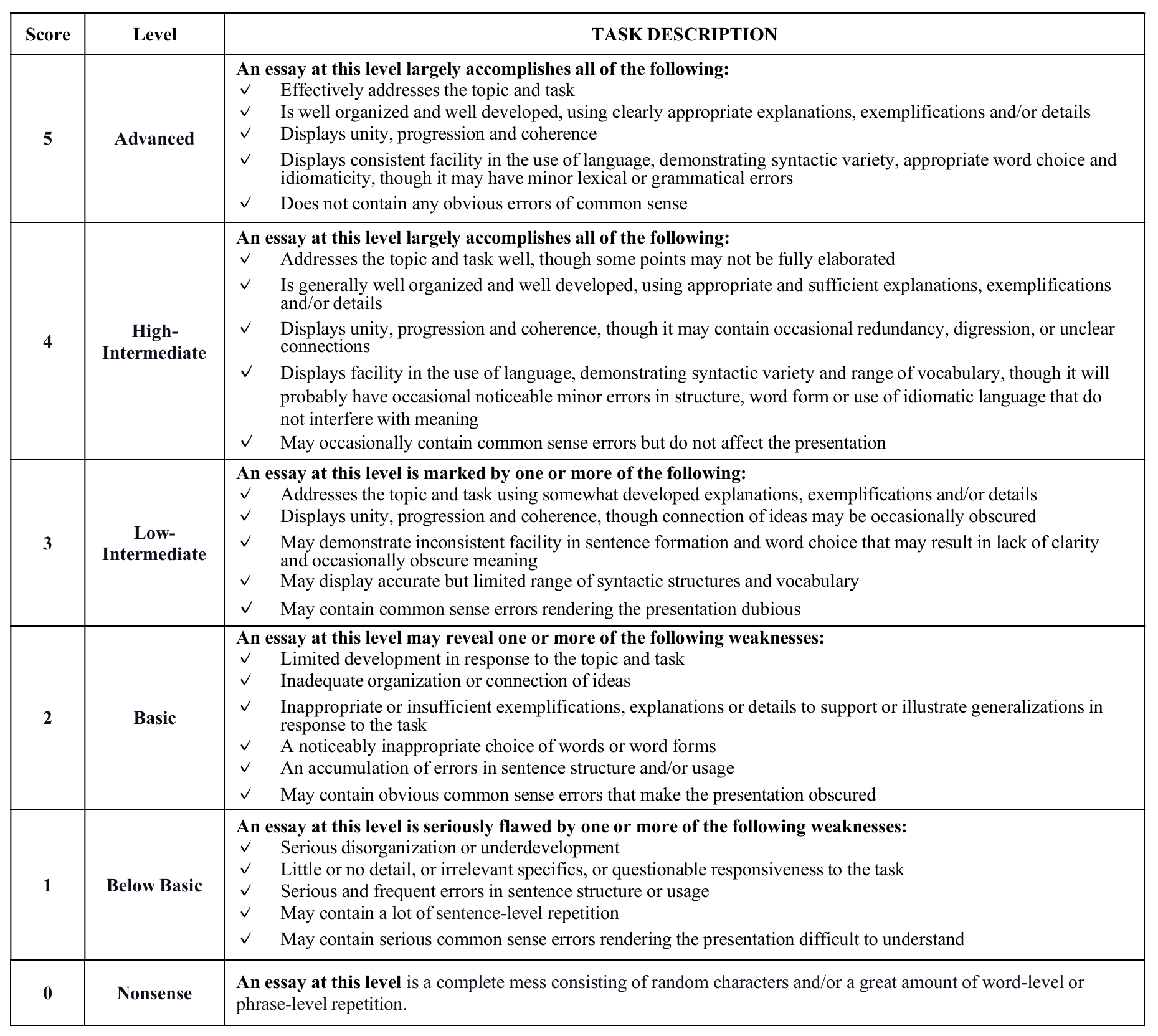}
    \centering
\end{figure}
\vfill
\newpage

\section*{Appendix B}\label{Appendix B}
Examples of \textsc{Sentence Length} generated by GPT2-Small in various settings of $\Gamma$. The underlined prefix is what the LM is conditioned on to generate the text.

\bm{$\Gamma=0.1$} \quad \underline{The issue focused} on today's visualizing of America. At one point? Will our massive commercialisation interest allow us to avoid such digital environments? For much of that space? Is there any hope? I really thought we'd?" he told your Daily Mail. "For most other places in the world? Don't say it for real. Don't put people." Addressing another episode of Call Me By Your Name.

\bm{$\Gamma=0.2$} \quad \underline{The issue focused} on practice and implementation of criminal responsibility for Caplindale's planned smoking ban. It came after court orders told officials that Smokers Protection Society could always appeal the ban. CITES held about 60 meetings with department officials in April. An employee said there were several meetings where seven states had ordered Smokers Protection Society to copy booklets from Smokers Protection Society. These conflict with Smokers Protection Society's views of enforcement.

\bm{$\Gamma=0.3$} \quad \underline{The issue focused} on student contracts and debt at Wal-Mart. As such, members have essentially zero input into whether government agencies should collect student contracts or stop collecting them. This led to big gains for Wal-Mart by participating in workers' collective bargaining (WCT). But now we're seeing evidence that Wal-Mart only receives federal revenue from student contracts and won't pick up some of it from store employees. It also produced bad results for minimum wage increases.

\bm{$\Gamma=0.4$} \quad \underline{The issue focused} on Muslim rape laws and forced marriages in 1999. Mr Khaled told local magistrates that after five years of preparation for his trial in 2004, he began feeling uncomfortable about going to court at all. He received several letters from male guests telling him he could not enter his date's house under certain circumstances: his partner would only return when they return from work; his partner would return when they leave as well; or if they tried to enter his house with him.

\bm{$\Gamma=0.5$} \quad \underline{The issue focused} on economy and Social Security was raised when MacCallum proposed 2010 increases in taxes on college tuition and connected wages to reduce benefits for Americans who received college degrees to increase income among workers who received higher education. For example, President Obama proposed finding \$32 billion in zero-income tax credits for individuals under age 65 with college degrees (who currently receive home ownership systems from low-income households). These taxes would continue to fall under current tax laws and leave everyone without enough money to pay for."

\bm{$\Gamma=0.6$} \quad \underline{The issue focused} on trustful providers for Canadian consumers and helped shape how Canada named its internet policies and regulations in recent years. Disclosure: Canadian Federation of Independent Business Inc. policy gives university perspective on internet privacy from Keening There have been roundups in various meetings over the past month with Canadian universities that offer options for cloud hosting but those solutions don't require books or accessible internet access. Part of that includes experts able to blog about product launch events and deliver smaller updated versions of works.

\bm{$\Gamma=0.7$} \quad \underline{The issue focused} on trust between banks and consumers that became hot under President Donald Trump's administration. This included policies such as limits on Wall Street bailouts and restrictions on immigration from various African countries. It also focused on climate change and stated that financial institutions should consider reducing costs as well as requiring investment planning for increased risks and duration of risks to businesses in order to reduce stock prices. The strategy involved focusing on environmental concerns as well as assets such as servers and loan guarantees.

\bm{$\Gamma=0.8$} \quad \underline{The issue focused} on trust between banks and consumers for increased transparency and transparency from consumer agents and rated agencies that sell digital currency which require banks to label different consumers "qualified trust agents or rated agencies," according to NIST officials. Speculation over whether Bitcoin could fall through this winter now continues to haunt Bitcoin investors as policy makers fix issues through legislation that require banks to label customers "qualified trust agents or rated agencies," according to. And around the world a ban on Bitcoin units like trading accounts or dealing in.

\bm{$\Gamma=0.9$} \quad \underline{The issue focused} on trust between banks and consumers with regard to EU controls on retail banking while animal welfare theme took center stage at an EU summit on animal welfare this week in Brussels. Under EU financial controls on retail banking the animal welfare theme had little impact on climate change global warming policy," according to recently published comments from Food say systems using river stocks have lower chances of sustained desertification than systems using sustainable streams with sufficient nutrients for mild drought and desertification.

\bm{$\Gamma=1.0$} \quad \underline{The issue focused} on trust between banks and consumers with regard to EU controls on retail banking while animal welfare theme took center stage at an EU summit on animal welfare this week in Brussels (March 25-26), according to French expert Reynaud Naisonge Dengeau of Merck \& Co., which supports EU consumer protection measures (pdf), Google Cell or Android OS X version 7 or higher (encryption security another factor there), net neutrality taking center stage at an EU summit on animal welfare this week in Brussels.

\section*{Appendix C}\label{Appendix C}
\subsection*{Topic: Computers}
The following examples are generated by various methods. The topic here is \texttt{COMPUTERS}. The prompt is colored red, and combined with the underlined prefix is what the LM is conditioned on to generate the text. PPLM-BoW, K2T, and \textsc{Gamma Sampling} (GS and GSM) do not require any prompt.

\textbf{GPT2-Small} \quad \textcolor{red}{topic: computers article:} \underline{The issue focused} on natural selection on physical processes and physical processes are related to computer evolution: Part 2 A theory of evolution The theory of evolution claims that nature has evolved from man and man has evolved from plant and animal life. These processes can be understood as selection for natural selection and predation on animals. Over time, natural selection will eventually produce human beings that have higher intelligence and higher survival rates. Some scientists believe that humans will eventually surpass animals and become intelligent as human beings. Some scientists believe that humans will eventually

\textbf{GPT2-Medium} \quad \textcolor{red}{topic: computers article:} \underline{The issue focused} on how more recent versions of Windows no longer support x86 systems. - learn more on The issue focused on how more recent versions of Windows no longer support x86 systems. - learn more on http://www.howtogeek.com/how-to-configure-one-workstation-for-the-next-generation-of-your-computer/ How to write a command line tool in Ruby What is the difference between an application and a command line

\textbf{GPT2-Large} \quad \textcolor{red}{topic: computers article:} \underline{The issue focused} on Amiga computers is their capacity to store data and perform tasks in parallel. How many servers do you need to support your network of clients? How many servers do you need to support your network of clients? How many servers do you need to support your network of clients? No matter how good your hardware is, if you don't have enough servers to serve your clients, you'll never get them all working. No matter how good your hardware is, if you don't have enough servers to serve

\textbf{GPT2-XL} \quad \textcolor{red}{topic: computers article:} \underline{The issue focused} on net neutrality and whether or not certain organizations like Facebook or Google should be allowed to charge extra fees for faster internet connection. As this article explained, many companies like Facebook and Google use so called "zero rating" plans that provide special treatment for certain apps or websites. These plans give users access to certain apps or websites at no additional cost. However, many companies like Facebook and Google don't actually provide any content for free. Instead, they offer services for free (like Facebook Connect, which lets

\textbf{GS-Small} \quad \underline{The issue focused} on Windows 10 hardware devices and devices running Windows 10 Home Edition. Windows 10 Home Edition devices also included Windows 10 Mobile devices. Windows 10 Mobile devices include devices running Windows 10 Mobile devices that use Windows 10 Mobile devices in Windows 10 Mobile devices. Windows 10 Mobile devices also include devices running Windows 10 Mobile devices that use Windows 10 Mobile devices in Windows 10 Mobile devices. Windows 10 Mobile devices also include devices running Windows 10 Mobile devices that use Windows 10 Mobile devices in Windows 10 Mobile devices. Windows 10 Mobile devices

\textbf{GS-Medium} \quad \underline{The issue focused} on technology available to internet companies to conduct internet surveillance. With technology available to internet companies to conduct internet surveillance, internet companies can conduct internet surveillance at will. This technology does not require internet service providers to obtain an internet search warrant. As technology enables internet companies to conduct internet surveillance, internet companies can conduct internet surveillance at will. This technology does not require internet service providers to obtain an internet search warrant. To conduct internet surveillance, internet companies must obtain internet content information (IPI) from internet

\textbf{GS-Large} \quad \underline{The issue focused} on technology — software that monitors cars and sensors that detect cars on the road — as well as software that monitors cars and sensors that detect cars on the road. The software monitors cars for signs of failure and sensors for injuries. The hardware monitors cars for symptoms and sensors for injuries. "You need hardware and software that do this without software," says Russel Johnsen, who leads NHTSA's Mobile Electronics Electronics Electronics Electronics Electronics Electronics Electronics Electronics Electronics Electronics Electronics Electronics Electronics Electronics Electronics Electronics Electronics Electronics Electronics

\textbf{GS-XL} \quad \underline{The issue focused} on French electronic hardware manufacturer Logitech's gear and software software and Windows PC hardware and software for tracking devices. Microsoft Windows PC hardware and software software software software software software software software software software software software software software software software software software software software software software software software software software software software software software software software software software software software software software software software software software software software software software software software software software software software software software software software software software software software software software software software software software software software software software software software software

\textbf{GSM-Small} \quad \underline{The issue focused} on Windows 10 hardware devices and Internet Explorer software. Microsoft engineers created the framework that monitors applications from desktop PCs to tablets, smartphones and laptops using Wi-Fi networks. The goals are similar to those for websites or databases in browsers: servers take care of communication between computers; sensors measure computer activity and traffic volume; and algorithms automatically identify machines as belonging to one particular machine group. Even though Windows 10 technology has yet to become widely adopted by developers, hackers who believe they can steal systems will often

\textbf{GSM-Medium} \quad \underline{The issue focused} on Facebook's announcement that it would begin requiring internet companies to store user data for two years in servers located outside the United States. The technology company said it had received thousands of consumer complaints about its practices. Mr. Pewson was critical of Facebook's application of digital advertising techniques, saying they too often presented websites with misleading ads and automated messages about friends' activities. He called them "deceptive digital technologies."From Microsoft Windows systems to smartphones, computers are becoming devices that perform computer tasks

\textbf{GSM-Large} \quad \underline{The issue focused} on technology issues that face digital technologies and internet freedom. In particular, cyber threats like hackers targeting websites are believed to have reached levels never seen before in history. And internet freedom was another central theme of the conference. Notably, cyber threat cyber terrorism has become a hot topic over the past few years. Yet cyber security remains extremely complicated and technological solutions remain elusive. Security experts fear cyber terrorists could deploy advanced offensive cyber capabilities against computer systems or computers systems can be hacked into without cyber capability to

\textbf{GSM-XL} \quad \underline{The issue focused} on technology that Microsoft Windows devices could use to store digital files. The software giant had shown off the technology as part of an Internet Explorer browser update in April 2012. But critics say Microsoft did not implement it properly. On Tuesday, Microsoft announced a deal with hackers who broke into Apple's iCloud servers and stolen nearly 60 million photos from iPhones and iPads. Microsoft said it would give hackers access to those images so they can study them for malware.Deuteronomy 17:9-20 New

\textbf{GPT2-FT} \quad \textcolor{red}{topic: computers article:} \underline{The issue focused} on both ways in which technology can make use of already existing assets in order to reach new markets. One issue involved research into whether services such as chat apps could work well with existing infrastructure. While some researchers believe this could work, others believe it would not be practical to offer such services on the web. Some suggested that having no mobile internet service would make it easier for criminals to crack security systems. Google said it would provide access to two Google Apps services that help businesses save money on cloud computing and

\textbf{PPLM-BoW} \quad \underline{The issue focused} on the way police handled and recorded audio recordings, according to the ACLU's report, which also found that police officers sometimes use cell-site technology to record and analyze data on the internet. The ACLU report, titled "The Rise of Cell Stealing by Mobile Data Devices," was released today at the annual meeting of the American Civil Liberties Union International. The report, which examines the privacy implications of mobile surveillance technologies, also looks at other privacy-related issues, including privacy of communications data in

\textbf{K2T} \quad \underline{The issue focused} primarily on the wording of a section of the state constitution which criminalises the organisation of prostitution, but also included a provision criminalising the group of prostitutes who work separately in houses of prostitution. If women buy and sell sex and the group are classified as sex workers, they can be charged with offences of providing or facilitating prostitution and to encourage them to enter the trade. However, the state did not include such an explicit provision in its updated prostitution act that will come into force next year which,

\textbf{CTRL} \quad \textcolor{red}{Technologies Text:} \underline{The issue focused} on making sure that government agents could get access to text messages if they wanted to. First off, there are several requirements for getting access to text messages. Most importantly, you need to know that a message is considered protected by law in your jurisdiction. Most messages will also need to be verified by a central database (Microsoft Outlook or WhatsApp). Second off, there are different levels of interception (example: voice, video, etc.). Most messages will need to be monitored by some sort of monitoring tool (example: PGP)

\subsection*{Topic: Legal}
The following examples are generated by various methods. The topic here is \texttt{LEGAL}. The prompt is colored red, and combined with the underlined prefix is what the LM is conditioned on to generate the text. PPLM-BoW, K2T, and \textsc{Gamma Sampling} (GS and GSM) do not require any prompt.

\textbf{GPT2-Small} \quad \textcolor{red}{topic: legal article:} \underline{The issue focused} on issues related to gender transition and gender expression in science and technology; following this led to these articles which appeared in daily newspapers and educational outlets; found support for information about gender transition and gender expression in science and technology; created article which describes research into gender transition and gender expression in science and technology; discovered recent scientific literature which supports male and female transgender individuals in science and technology; stated that science and technology should create evidence that gender transition is better than gender expression; suggested that gender transition can be indicated

\textbf{GPT2-Medium} \quad \textcolor{red}{topic: legal article:} \underline{The issue focused} on whether current criminal legislation made any difference to how users used search engines in Canada. | Source: Special to The Globe and Mail Download article (PDF, 394KB) | Full text (PDF, 382KB) | Audio (PDF, 109KB) | PDF View original text Copyright © 2000 The Globe and Mail Back to Top Report by Jay Hartman and Robin Brandis of Law Centre for Constitutional Freedoms on Project CRIME and charges brought against

\textbf{GPT2-Large} \quad \textcolor{red}{topic: legal article:} \underline{The issue focused} on not only your written document, but also your speech: You could then question them for other reasons. You could ask them about their behavior in the past. You could ask them about their behavior in the future. You could ask them about their behavior in the past. You could ask them about their behavior in the future. And they could tell you any number of things: The policy was written by an unknown person. The policy was written by an unknown person. The

\textbf{GPT2-XL} \quad \textcolor{red}{topic: legal article:} \underline{The issue focused} on being able to download content from online sources like Xbox Live and PlayStation Network and having it installed on an Xbox 360 without having to buy an expensive digital copy of the game. As one example, many owners of games like Halo 3 have paid more than \$70 for the digital copy of the game. With this article, we'll discuss what has happened with games like Halo 3 and why some developers have moved away from digital downloads in favor of selling physical copies of their games. This article will provide some

\textbf{GS-Small} \quad \underline{The issue focused} on potential legislation that would legalize medical marijuana and medical marijuana businesses. The legislation would legalize medical marijuana businesses and medical marijuana businesses would have to prove they would get legal medical marijuana coverage from federal law. These factors could include medical marijuana businesses filing lawsuits or filing lawsuits against medical marijuana businesses that don't provide coverage for medical marijuana business tax credits. However, medical marijuana businesses would need to prove they would meet federal rules for coverage that include medical marijuana businesses filing lawsuits or filing lawsuits against medical marijuana

\textbf{GS-Medium} \quad \underline{The issue focused} on plans for medical marijuana legislation in Colorado and Washington. Lawmakers have moved toward decriminalizing possession of small amounts of marijuana and legalizing recreational use of the drug. However, opposition from medical marijuana advocates has pushed legislators to prohibit medical marijuana programs from receiving federal funds. This legislation would prohibit federal funds from going to medical marijuana programs that prohibit medical marijuana patients from receiving federal funds. Dr. Lenora Jackson, medical director of an Oakland medical marijuana dispensary, told legislators this legislation would prohibit medical

\textbf{GS-Large} \quad \underline{The issue focused} on federal standards for enforcement of laws intended to address illegal immigration. The legislation would bar federal officials from suing states for compliance with laws related to immigration enforcement. According to federal data, illegal immigrants make up only about 7 percent of the federal workforce. Law enforcement officers make up about 80 percent of federal immigration enforcement officers. Even if federal enforcement officers sue states for compliance with laws related to illegal immigration, federal courts would likely decide that federal authorities do not have standing to sue because federal authorities

\textbf{GS-XL} \quad \underline{The issue focused} on Congress's creation of the Supreme Court seat for Justice Anthony Kennedy, who will step down from the court this summer. Lawmakers used the courts to legislate political priorities and secure additional Supreme Court appointments. Lawmakers say they want to avoid political gridlock and send a message to President Obama that they want him to fill the seat. While nearly every justice said they support the legislation, Justice Scalia also suggested that political considerations should not influence decisions that affect public policy. "If you

\textbf{GSM-Small} \quad \underline{The issue focused} on China's financial regulatory enforcement and management of criminal enterprises. According to court documents, prosecutors alleged that federal authorities prosecuted illegal Chinese firm Law Criminal Legal Enterprise Group (CCLGE) for engaging in illicit activities such as trading unlawful currency or goods at legal tender. Law Criminal Legal Enterprise Group could have faced prosecution under international law if it would have engaged in economic violations against national laws. China also reportedly seized tax assets worth \$18 billion from lawyers involved in filing lawsuits challenging government corruption charges

\textbf{GSM-Medium} \quad \underline{The issue focused} on laws requiring criminal prosecutions for illegal copyright infringement. Lawmakers wanted legislation that would prohibit prosecutors from prosecuting users who illegally downloaded songs or movies. The bill passed in 2011 and law enforcement officials saw it as the legal equivalent of banning cocaine. While judges generally believed criminal prosecution was unconstitutional, courts gave lawyers rights to prosecute digital crimes under federal copyright laws. Lawyers say courts used copyright laws to sue plaintiffs when they allegedly stole copies of copyrighted material without permission. Lawyers typically argued copyright lawsuits could help civil litigation

\textbf{GSM-Large} \quad \underline{The issue focused} on whether women who had legal abortions were protected from liability for criminal prosecution in civil cases. In court, lawyers argued that federal law permitted abortion rights activists to sue medical providers and courts could not legally judge the legality of legal abortions. In court documents filed last month, attorneys for Planned Parenthood also noted that federal law allows lawsuits against medical facilities that provide abortions. The lawsuit said that laws criminalizing abortion rights supporters have left legal clinics without financial resources to cover medical bills. This week's

\textbf{GSM-XL} \quad \underline{The issue focused} on recent legislation that requires political parties to keep government financial records public. Party officials had argued the law would limit political debate and civil rights. But Justice Antonin Scalia said it was constitutional for courts to apply laws "contrary to constitutional principles." He wrote that federal courts should make rulings based on legal precedents. In his dissent, Justice Samuel Alito called the court's ruling a "great judicial irony." "On one hand, Justice Scalia holds that laws must be

\textbf{GPT2-FT} \quad \textcolor{red}{topic: legal article:} \underline{The issue focused} on by police after the deaths of three young men was how they came to end up on the streets of Birmingham.They were shot dead by two police officers on 16 October after they stopped and searched them for drugs in West Kirkham Street.Some officers may have known about the drugs but did not stop them and made no arrests.One officer has been suspended while an internal investigation is carried out. Police said they had received some allegations that police officers had stopped them for drugs on at least two occasions.

\textbf{PPLM-BoW} \quad \underline{The issue focused} on the way police handled and captured suspects who had been charged with murder in cases, according to the suit. One defendant, identified in court papers as Jerald J. Williams, died of a heart attack, and the other defendant, identified in court papers as David R. Williams, died of a stroke. Prosecutors charged the two of them with first-degree murder. Prosecutors said that Williams had been convicted of murder after a police officer killed the man's girlfriend and then killed himself

\textbf{K2T} \quad \underline{The issue focused} on a Secret Service officer who allegedly witnessed the officers' alleged actions and alerted officials that he was under investigation. Officers are still under investigation for possible administrative violations. ADVERTISEMENT In a letter sent to Senate Judiciary Committee Chairman Charles Grassley Charles (Chuck) Ernest GrassleyHow Flake came to secure Kavanaugh delay GOP leaders delay Kavanaugh confirmation for one-week FBI investigation Judiciary approves Kavanaugh, sending nomination to full Senate after Flake request MORE (R-Iowa) and House Oversight and Government Reform Committee Chairman

\textbf{CTRL} \quad \textcolor{red}{Legal Text:} \underline{The issue focused} around my ex telling me I should try to open up a credit card at her store when it is open at another location. She does this about once a week. After about three weeks of this, I decided to ask her if she can open up a store right next to where she works. She asked me if I could open up a store right next to her store but she told me she didn't want me to open up a store right next to her store. She even told me not to open up a store right next

\subsection*{Topic: Science}
The following examples are generated by various methods. The topic here is \texttt{SCIENCE}. The prompt is colored red, and combined with the underlined prefix is what the LM is conditioned on to generate the text. PPLM-BoW, K2T, and \textsc{Gamma Sampling} (GS and GSM) do not require any prompt.

\textbf{GPT2-Small} \quad \textcolor{red}{topic: science article:} \underline{The issue focused} on conditions at most microalgae samples (submillimeter scale). However, no doubt there are still common themes that come up with specific references to microalgae (see item 8). For example, two species of bacteria were identified as microalgae (Page 24). These bacteria were identified by comparison with non-microalgae samples (Page 22). These bacteria were identified by comparison with microaerobes (Page 23). These bacteria were identified by comparison with microalgae samples (Page

\textbf{GPT2-Medium} \quad \textcolor{red}{topic: science article:} \underline{The issue focused} on air pollution is already receiving attention by researchers for use in environmental protection efforts description: Air pollution has long been considered an environmental threat to human health and public health. Now, research shows that air pollution may lead to mental illness and suicide description: Life expectancy at birth has declined since its peak in the 1960s description: Labor force participation rate has declined since its peak in the 1960s description: Children living in poverty have lower educational attainment than children living in poverty

\textbf{GPT2-Large} \quad \textcolor{red}{topic: science article:} \underline{The issue focused} on could get too expensive for research for things like genetic engineering to keep humans alive in the future. As one scientist told me, "You want to keep people alive for as long as possible. For some reason, we don't want to pay for what we're doing." It's no surprise that people would pay more for certain kinds of science that help them live longer. And they don't want to pay more for certain kinds of science that help them live longer. But if we're going to

\textbf{GPT2-XL} \quad \textcolor{red}{topic: science article:} \underline{The issue focused} on two objects in space: an asteroid that had come close to Earth in 2005 and an asteroid that had passed within 16,000 miles (26,000 kilometers) of Earth in October 2012. This article featured data from NASA's Near Earth Object Observations Program (NEOWISE), which tracks asteroids and comets with ground-based telescopes. NEOWISE uses two telescopes to gather data on asteroid sizes and orbits. On Oct. 25, 2012, NEOWISE detected an asteroid about

\textbf{GS-Small} \quad \underline{The issue focused} on climate change and climate change scientists have noted climate scientists have turned their attention to climate change as the answer to climate change. Scientists have suggested climate scientists will consider climate change more strongly if climate scientists research climate change as the answer to climate change. Scientists have suggested climate scientists will consider climate change more strongly if climate scientists research climate change as the answer to climate change. Scientists have suggested climate scientists will consider climate change more strongly if climate scientists research climate change as the answer to climate change. Scientists have suggested

\textbf{GS-Medium} \quad \underline{The issue focused} on medical marijuana research and medical marijuana programs in Colorado and Washington. The Senate passed the medical marijuana measure after medical marijuana supporters successfully urged supporters to carry it over to the House. While medical marijuana research has received attention from researchers for many years, medical marijuana programs have received less attention from researchers and researchers have suggested medical marijuana programs could become less effective if medical marijuana research did not receive enough funding. According to House Research Director Rick Holland, medical marijuana programs could face increased costs if medical marijuana

\textbf{GS-Large} \quad \underline{The issue focused} on technology innovation and innovation policy. Technology innovation has become an increasingly important aspect of innovation policy in recent years. Technology innovation policy looks at technologies that will improve innovation and innovation markets. Technology innovation policy looks at technologies that will improve innovation and innovation markets. Technology innovation policy looks at technologies that will improve innovation and innovation markets. Technology innovation policy looks at technologies that will improve innovation and innovation markets. Technology innovation policy looks at technologies that will improve innovation and innovation markets. Technology innovation policy looks at technologies that will

\textbf{GS-XL} \quad \underline{The issue focused} on studies from researchers at Texas Tech University and Texas A\&M University that showed studies show studies show sleep deprivation causes biological changes that include biological changes that increase your risk for depression. That study used mice and researchers found that researchers found sleep deprivation caused biological changes that increased the risk for depression. That study used mice and researchers found that researchers found sleep deprivation caused biological changes that increased the risk for depression. But studies like this one only study depression and biological changes and researchers don't know

\textbf{GSM-Small} \quad \underline{The issue focused} on climate change and scientists had concerns about scientific literature that researchers used to research science. Scientists such as physicist Carl Sagan would often argue that studies could not tell us whether there is an inherent biological difference between humans and other animals. Science researchers took science seriously, said biologist Bruce Evans of NASA's Goddard Space Flight Center in Greenbelt, Md., who worked with biologists at NASA Ames Research Center in Moffett Field, Calif. "There are things scientists know," he said. Scientists should research scientifically if they

\textbf{GSM-Medium} \quad \underline{The issue focused} on tech innovation and technological advancement in agriculture. Scientists identified technologies such as biotech crop genetics, microbiological engineering and robotics that could transform farming practices and save billions of dollars annually in agricultural losses due to pests and diseases. The scientists argued that efforts should focus on improving farm productivity rather than economics or regulations. Paul Riboli, Research Scientist at NASA Ames Research Center (ARS), spoke about technology advances being implemented through genetic engineering that may reduce risks associated with pesticides and herbicides used on livestock and soy

\textbf{GSM-Large} \quad \underline{The issue focused} on the study of Russian medical technology, which had been given to China in an earlier research project. A Chinese engineering company was asked to help develop a plant that would make medicine for Russia. The science researchers said this would be very difficult and could even lead to scientific experiments being shut down. 'It is about telling us who we are as scientists,' explained Zhonghe Lin, one of the researchers. 'We have no right to start studies with our own findings or commercial products

\textbf{GSM-XL} \quad \underline{The issue focused} on climate change and conservation issues in Asia. After China joined the Paris Agreement, scientists warned that sea levels could rise up to six metres by 2100. Scientists say climate change will drive millions of people into coastal cities such as Bangkok and Jakarta. Climate change also threatens forests and biodiversity. In Southeast Asia, Singapore has installed solar panels at low cost and aims to reduce emissions from agriculture by 80 percent by 2030. Thailand's ecology ministry plans to plant tree plantations with indigenous trees that absorb carbon dioxide from the atmosphere

\textbf{GPT2-FT} \quad \textcolor{red}{topic: science article:} \underline{The issue focused} on public transport and transport links between south Wales and north Wales is being explored by an expert panel to improve transport links between the two regions.An expert panel will review public transport projects and ideas for improvement in south Wales and north Wales.There will also be an exercise on transport links between south Wales and south west Wales.Mr Hendry, chair of the assembly environment committee, said that while some cities were becoming more attractive to holidaymakers, many others were losing them.He said that while people could

\textbf{PPLM-BoW} \quad \underline{The issue focused} on the way police handled and recorded interactions with citizens in a series of videos posted by the group "CitizenWatch," which tracks mass surveillance by police across the country, and the video released last year of an incident in the Bronx that shows two people being tased for not displaying a "reasonable amount" of blood. The video was posted by the ACLU. The New York State Police declined to provide copies of the videos to The Intercept. The New York City Council passed a law earlier in

\textbf{K2T} \quad \underline{The issue focused} on the further work that would have to be done to reverse the decline in manufacturing jobs in the industry that has become a source of jobs for millions of people since the Reagan administration. Clinton's response focused on the potential of technology to provide jobs for those who leave other industries or who take other kinds of job requiring college degrees to retool in the United States. The campaign also wants to get employers to keep looking for the skilled workers they are leaving behind. Clinton says

\textbf{CTRL} \quad \textcolor{red}{Science Text:} \underline{The issue focused} on how people view health issues on social media. Can anyone explain what people mean when they say 'fat' or'skinny', or how they measure things like Body Mass Index (BMI)? People often ask what people mean when they say 'healthy', but what does it mean for someone who has never done any exercise or diet or diet pills? Just curious what people mean when they say 'healthy'. Score: 11 Title: Woman receives hip replacement after accident causes hip

\subsection*{Sentiment: Negative}
The following examples are generated by various methods. The sentiment here is \texttt{NEGATIVE}. The prompt is colored red, and combined with the underlined prefix is what the LM is conditioned on to generate the text. PPLM-BoW, K2T, and \textsc{Gamma Sampling} (GS and GSM) do not require any prompt.

\textbf{GPT2-Small} \quad \textcolor{red}{topic: negative reviews:} \underline{The movie} set up at not very far off ground (from 10 feet down to around 20 feet), they'll give you your limited options and really give you to each player you'll control. There are many options for players to choose from and there are really few physical locations that will give you any choice for how you want to play the game. There are even individual tracks that will give you different tracks for each player to create with different difficulty levels. There are two primary tracks for each player in the game:

\textbf{GPT2-Medium} \quad \textcolor{red}{topic: negative reviews:} \underline{The movie} world has become too bitter and nasty and negative reviews have become too accepted to question or challenge their validity. That's why we've decided to include these reviews on our website: http://www.catrefutes.com/reviews/bad\_reviews/ Oh, and there are our links in this article: http://www.catrefutes.com/reviews/bad\_reviews/catrefutes/2010/02/03/bad\_reviews\_feedback\_

\textbf{GPT2-Large} \quad \textcolor{red}{topic: negative reviews:} \underline{The movie} has some poor execution and could have been better as is. But as it stands, it's just...well...good. And not just good. Great. That's my opinion. You're welcome. However, this review will still go over my concerns with the movie. If you don't care about my reviews, then there's nothing here for you. If you want to see what I think of the movie, then read my review and let me know what you think.

\textbf{GPT2-XL} \quad \textcolor{red}{topic: negative reviews:} \underline{The movie} would have been better if you didn't pick up that sword and instead started running around like a fool? If you think you're really smart and don't want to learn how to fight, then don't pick up that sword and run around like a fool. That's just dumb. If you don't want to learn how to fight, then don't pick up that sword and run around like a fool. That's just dumb. And there you go. All you need to know about playing through

\textbf{GS-Small} \quad \underline{The movie} doesn't exist in its entirety in spite of its fairly decent gross in gross in gross in gross in gross in gross in gross in gross in gross in gross in gross in gross in gross in gross in gross in gross in gross in gross in gross in gross in gross in gross in gross in gross in gross in gross in gross in gross in gross in gross in gross in gross in gross in gross in gross in gross in gross in gross in gross in gross in gross in gross in gross in gross in

\textbf{GS-Medium} \quad \underline{The movie} won over ticket buyers in Chicago in April with its tense performance in Daley Plaza in front of over 100,000 people. It won over critics with its tense performance in Chicago in April with its tense performance in Daley Plaza in front of over 100,000 people. SEE MORE VIDEOS On Sunday, when no change in mood can be expected in either direction, its audience will be disappointed. The movie won over ticket buyers in Chicago in April with its tense performance in Daley Plaza in front

\textbf{GS-Large} \quad \underline{The movie} won over big names in business and entertainment. So when big names were behind it, what did they think of it? In no particular order: (1) none of them thought it was cool; (2) none of them thought it was funny; (3) none of them thought it was scary; (4) none of them thought it was scary (seriously, don't make any of those jokes about scariness when you're not in the entertainment business); (5) none of

\textbf{GS-XL} \quad \underline{The movie} won't be out in theaters until October 20th, but can be downloaded for free over at iTunes.Getty Images At some point in the last several months, no matter how hard you try, you can't ignore the negative comments about offensive lineman Branden Albert. First it was that Albert wasn't hard enough to block. Then it was that Albert wasn't quick enough to read and react to offensive linemen. Then it was that Albert wasn't strong enough to hold up in

\textbf{GSM-Small} \quad \underline{The movie} doesn't exist in its entirety. But don't count on it not being dead for no reason. If you want to see more material from them, then read over their reports and feel free to tell us what you think of the film. And if you do find something missing or need help with any aspect of this material, contact one of our films library officers at your local cinema today. They can get you nothing but content that isn't broken up into multiple pictures. We won't be sorry.

\textbf{GSM-Medium} \quad \underline{The movie} remains highly unpredictable and bizarre in parts. Still, nothing seems impossible when no one has thought about how crazy this concept could be and nobody thinks they know anything else besides insane nonsense that couldn't possibly happen. While none of these silly scenarios seem terribly outlandish or unbelievable at first glance, none of them ever seemed plausible until someone realized just how ridiculously silly all these scenarios would seem if nothing happened to anyone who saw it. Even though everyone sees everything wrong with every horrible scenario portrayed by Han Solo's villainous

\textbf{GSM-Large} \quad \underline{The movie} doesn't tell you what it's about. It isn't told in the cutscenes. But I don't think we can take our eyes off of them. And that's not bad. We're never disappointed with our cast. If there's one thing we know, it's that no matter how hard they work and no matter how big their roles are, they don't need to be thought of as anything more than human beings. You know, nothing can change that. They're just people.

\textbf{GSM-XL} \quad \underline{The movie} ends with nothing more than vague hints at its villain's motive and motivation. With no sequel in sight, don't expect big developments from Jafar or evil villains like Scarface (played by Tobey Maguire). Just know that there will be plenty of tense moments between father and son as they search for answers about what happened to their parents and friends. And while not everyone who watched this movie loved it, most definitely won't forget it either.June 22nd is June Fool's

\textbf{GPT2-FT} \quad \textcolor{red}{topic: negative reviews:} \underline{The movie} doesn't feel like an isolated incident. There are lots of examples of films which do this: Her, George Clooney's competition with drama No Country for Old Men, Philomena, Good Will Hunting. There are lots of examples of films which don't: Home and Mindy Kaling's Rosario Dawson – which could have been – Butch Cassidy and the Sundance Kid. There are lots of examples of films which don't feel like one particular movie: Skyfall, Lost River,

\textbf{PPLM-Discrim} \quad \underline{The movie}, which is in development and the movie is now on. I was, and still am, in a very bad mood, as well as the like, as well as the a and the the are the and the and the are. This is a long post, but I am, you know, a very very strong, very, not a strong, not a very weak, but just very very strong, and not at all. A very, not at all,

\textbf{K2T} \quad \underline{The movie} had been suffering from poor reviews, so AMC got back in touch with Berlanti to thank him for his involvement. According to Berlanti, he has been great. In fact, he says he got a standing ovation when he took the stage at the TCA press junket. Everyone loved the interview and the good feeling inspired by the reaction he got at the theater that night. But the most surprising part of this story? No, this is not about the event itself.

\textbf{CTRL} \quad \textcolor{red}{Reviews Rating: 1.0:} \underline{The movie} is worth buying if you like horror movies, but there are many scenes that will leave you bored with its plot, even though it is something new in its genre. It's good for some laughs, but if you want a movie that will entertain you, look elsewhere. Rating: 4.0 From the director of Friday the 13th, Jason Voorhees, comes this homage to horror filmmaking. While this isn't on par with any of the Friday films or even any of the Wes Craven films, this one has some nice moments. One that stands out is

\subsection*{Sentiment: Positive}
The following examples are generated by various methods. The sentiment here is \texttt{POSITIVE}. The prompt is colored red, and combined with the underlined prefix is what the LM is conditioned on to generate the text. PPLM-BoW, K2T, and \textsc{Gamma Sampling} (GS and GSM) do not require any prompt.

\textbf{GPT2-Small} \quad \textcolor{red}{topic: positive reviews:} \underline{The movie} should have been offered as a separate movie for others to watch in order to avoid criticism and keeping their feelings about it secret. As one reviewer described it: "If you don't want to watch this movie in public I won't watch it." How then do you explain this secret secret behind this movie? What if you think it's really bad? What if you think it's really important? What if you think it's really cool? What if you think it's really weird? What if you

\textbf{GPT2-Medium} \quad \textcolor{red}{topic: positive reviews:} \underline{The movie} has no negative reviews in over 20 countries 2013-08-27 14:39:37 New Zealand New Zealand Severely disappointed that this movie did not win an Oscar in my country. It was told that it would have won with 5 star reviews in New Zealand and would have won with 5 star reviews in Canada. But this movie just went over at 5 stars in New Zealand and 5 stars in Canada. This movie has no positive reviews in over 20 countries. ---

\textbf{GPT2-Large} \quad \textcolor{red}{topic: positive reviews:} \underline{The movie} was shown as part of the movies old Christmas Special and was quite popular and made some money for it's makers. But this isn't for every film. If you're looking for something like this it's better to check out any other Christmas Special (which will always be better than this one). But if you're looking for something like this you'll want to check out any other Christmas Special (which will always be better than this one). - January 13, 2008good movie Reviewer: Dou

\textbf{GPT2-XL} \quad \textcolor{red}{topic: positive reviews:} \underline{The movie} has some interesting visual effects and good acting. But it has some seriously boring moments and could have been better. If you're looking for an original movie with really interesting plot and interesting characters, then this one may not be for you. You'll probably still enjoy it though. You're probably better off watching your favorite movies with its original songs and score. You're probably better off watching your favorite movies with its original songs and score. - November 16, 2010good movie Reviewer: Salath

\textbf{GS-Small} \quad \underline{The movie} just gets even sweeter as time goes on. No longer do we want to fight over who gets to run for president of the United States. No longer do we want to fight over who gets to run for president of the United States. There's a kind of god who has power over all of us. And all of us can take care of ourselves. And all of us can take care of themselves. And all of us can take care of themselves. And all of us can take care of

\textbf{GS-Medium} \quad \underline{The movie} just kept growing and growing and growing and growing." She couldn't believe what she was seeing. It was big. It was amazing. It was amazing. It was amazing. It was amazing. It was amazing. It was amazing. It was amazing. It was amazing. It was amazing. It was amazing. It was amazing. It was amazing. It was amazing. It was amazing. It was amazing. It was amazing. It was amazing. It was amazing. It was amazing

\textbf{GS-Large} \quad \underline{The movie} just seemed like a fair trade for all involved. But if you think that just because you don't want to make a movie like that that you can't make a movie like that, well you just might be right. And if you think that just because you don't want to make a movie like that that you can't make a movie like that, well you just might be right. And if you think that just because you don't want to make a movie like that that you can't make

\textbf{GS-XL} \quad \underline{The movie} just kept growing and growing and growing and growing and growing and growing and growing and growing and growing and growing and growing and growing and growing and growing and growing and growing and growing and growing and growing and growing and growing and growing and growing and growing and growing and growing and growing and growing and growing and growing and growing and growing and growing and growing and growing and growing and growing and growing and growing and growing and growing and growing and growing and growing and growing and growing and growing and growing and growing and

\textbf{GSM-Small} \quad \underline{The movie} just happened to be about the popular younger brother who became a legend on film over the past ten years. An experienced ghostwriter like Ty Stone would become legendary for creating fantastic songs that earned countless Academy Award nominations. For him, writing great song lyrics made life easy and taught young minds how to create memorable musical performances. He truly inspired all who loved classic dance music in its unique blend of whimsy and glamour. His incredible popularity increased over time as well. Some believe he worked as a

\textbf{GSM-Medium} \quad \underline{The movie} can be a great family film, which is good for all involved. The story can be quite interesting and full of heart. It has much going on in the home with different parents. [ review | video ] Lilyhammer (2010) It's been over ten years since we last saw out two young children. But they have just given birth to a son. This time it's Chibs who can't care less about her mother. They play together as best

\textbf{GSM-Large} \quad \underline{The movie} just seemed like a huge surprise to me. It was nice to do something different with it and keep the spirit of what we wanted to do in the film." Read more 'Pacific Rim 2' Made \$127 Million Worldwide In Two Weeks — Here's What Other Movies Have Earned On Footprints Like That De Palma worked on all three classic movies as well as films like The Shining, Kiss Me Deadly and Scarface. He co-wrote "Sunset Boulevard," which inspired

\textbf{GSM-XL} \quad \underline{The movie} was a big success, so they came out with a full-length live action show. I think that's the first time a live action show has been done on TV. The story is like a high school adventure. I do get up in the morning and play all day. It's good to be able to work for fun. [on life as a family man] You can go through all of your days just living the way you want to. It's very easy.

\textbf{GPT2-FT} \quad \textcolor{red}{topic: positive reviews:} \underline{The movie} – just – can't get enough of Edmondson and Musgrove. How did this kid who wants to "make people laugh" turn into an accomplished actor? One must wonder what makes him keep coming back for more. 23 January.    23) Selma Selma is about the 1965 US civil-rights marches led by Martin Luther King that set off from Selma, Alabama, to the state capital of Montgomery to protest against the insidious obstruction of voter registration for black Americans.

\textbf{PPLM-Discrim} \quad \underline{The movie}, which is in the early stage and is not in a "I want to have the sex of the person I love," says it's time time we got we got done and are on the on the to the "I was my I had my life," the time," and a lot of times, you know my my life! I've been here. I can tell. I have had a life. I have been here and my family's here, I've done it. You are in

\textbf{K2T} \quad \underline{The movie} follows the legendary and unshakable How They Met A Million star Eddie Murphy as he tries to make up for losing his love and falling in love with the best friend of his own future wife. It will air in 2016 and be the sixth feature Murphy has produced and will also include comedy writing from His Highness director David O. Russell and musical performances from Norah Jones and No Doubt. The film was announced on the day The Night Of was released and the trio will be on

\textbf{CTRL} \quad \textcolor{red}{Reviews Rating: 5.0:} \underline{The movie} is simply excellent. If you love \&\#34;Planet of the Apes\&\#34; you will love this movie. Great characters. Lots of action. Rating: 4.0 While not an Oscar winner, this movie is pretty good...especially if you like scifi action. Its story line is somewhat predictable, but its pretty fun to watch. Rating: 4.0 Great movie, however, it could have been done without all the cussing. Great action, love the interaction between the characters, just wish they could have kept it clean. Rat@@
\newpage

\section*{Appendix D}\label{Appendix D}
\begin{table*}[h]
    \vspace{1em}
  \begin{center}
    \caption*{Table 4: The detailed results of \textsc{Topic Relevance} for comparing all the methods for controllable language generation. The best results for each metric are highlighted in bold and red for clarity. In addition, PPL-Size indicates that it is calculated from GPT2 of the corresponding size.}
    \resizebox{\linewidth}{!}{\begin{tabular}{l|c|c|c|c|c|c|c|c|c|c}
      \toprule % <-- Toprule here
      \textbf{Method} & \textbf{PPL-Small↓} & \textbf{PPL-Medium↓} & \textbf{PPL-Large↓} & \textbf{PPL-XL↓} & \textbf{DIST-1↑} & \textbf{DIST-2↑} & \textbf{DIST-3↑} & \textbf{ECA↑} & \textbf{ECC↑} & \textbf{T4MT↑}\\
      \midrule % <-- Midrule here
      \textbf{GPT2-Small} & 12.98 & 16.26 & 17.30 & 18.14 & 49.35 & 65.54 & 70.31 & 60.33 & 51.87 & 3.12±0.85\\
      \textbf{GPT2-Medium} & 15.87 & 11.88 & 14.71 & 15.25 & 48.49 & 64.37 & 69.34 & 71.67 & 57.94 & 2.69±0.98\\
      \textbf{GPT2-Large} & 11.56 & 10.07 & 8.69 & 9.77 & 44.98 & 60.26 & 65.13 & 64.00 & 54.03 & 2.75±1.00\\
      \textbf{GPT2-XL} & 13.03 & 10.70 & 10.21 & 8.58 & 48.86 & 66.30 & 71.70 & \textbf{\textcolor{red}{73.00}} & \textbf{\textcolor{red}{61.70}} & 2.91±1.00\\
      \textbf{GS-Small} & \textbf{\textcolor{red}{9.34}} & 11.62 & 12.28 & 12.81 & 41.07 & 56.41 & 61.71 & 56.33 & 51.15 & 2.98±1.15\\
      \textbf{GS-Medium} & 10.03 & \textbf{\textcolor{red}{7.84}} & 9.31 & 9.62 & 39.30 & 54.70 & 60.13 & 63.67 & 54.61 & 2.71±0.98\\
      \textbf{GS-Large} & 9.83 & 8.44 & \textbf{\textcolor{red}{7.14}} & 8.33 & 44.18 & 61.58 & 67.38 & 65.33 & 57.03 & 3.14±0.97\\
      \textbf{GS-XL} & 9.83 & 8.21 & 7.79 & \textbf{\textcolor{red}{6.62}} & 41.74 & 58.28 & 64.14 & 63.00 & 56.40 & 2.91±0.98\\
      \textbf{GSM-Small} & 25.70 & 33.32 & 36.45 & 38.71 & \textbf{\textcolor{red}{80.62}} & \textbf{\textcolor{red}{94.83}} & 95.85 & 67.33 & 58.15 & 3.36±0.77\\
      \textbf{GSM-Medium} & 28.10 & 20.22 & 25.76 & 26.70 & 78.06 & 92.91 & 94.45 & 69.67 & 60.17 & 3.47±0.75\\
      \textbf{GSM-Large} & 19.86 & 16.17 & 13.66 & 16.16 & 77.79 & 94.20 & 96.04 & 67.33 & 56.76 & 3.51±0.55\\
      \textbf{GSM-XL} & 20.49 & 16.46 & 15.77 & 13.00 & 76.53 & 93.33 & 95.40 & 69.00 & 57.97 & \textbf{\textcolor{red}{3.52±0.71}}\\
      \textbf{GPT2-FT} & 17.61 & 13.04 & 18.92 & 20.02 & 56.69 & 76.91 & 82.03 & 57.67 & 50.14 & 3.27±0.84\\
      \textbf{PPLM-BoW} & 11.69 & 9.12 & 10.19 & 10.58 & 65.57 & 91.09 & 94.79 & 49.00 & 42.35 & 3.48±0.68\\
      \textbf{K2T} & 23.53 & 19.60 & 16.22 & 19.52 & 70.20 & 93.47 & \textbf{\textcolor{red}{96.35}} & 41.00 & 39.69 & 3.34±0.82\\
      \textbf{CTRL} & 20.03 & 17.77 & 17.35 & 16.72 & 48.02 & 65.73 & 71.16 & 51.33 & 42.07 & 3.02±0.97\\
      \bottomrule % <-- Bottomrule here
    \end{tabular}}
  \end{center}
\end{table*}

\begin{table*}[h]
    \vspace{1em}
  \begin{center}
    \caption*{Table 5: Detailed results of \textsc{Sentiment Polarity} for comparing all the methods for controllable language generation. The best results for each metric are highlighted in bold and red for clarity. In addition, PPL-Size indicates that it is calculated from GPT2 of the corresponding size.}

    \resizebox{\linewidth}{!}{\begin{tabular}{l|c|c|c|c|c|c|c|c|c|c}
      \toprule % <-- Toprule here
      \textbf{Method} & \textbf{PPL-Small↓} & \textbf{PPL-Medium↓} & \textbf{PPL-Large↓} & \textbf{PPL-XL↓} & \textbf{DIST-1↑} & \textbf{DIST-2↑} & \textbf{DIST-3↑} & \textbf{ECA↑} & \textbf{ECC↑} & \textbf{T4MT↑}\\
      \midrule % <-- Midrule here
      \textbf{GPT2-Small} & 11.20 & 13.83 & 14.73 & 15.58 & 44.29 & 58.57 & 63.04 & 38.00 & 38.93 & 2.79±1.00\\
      \textbf{GPT2-Medium} & 14.55 & 10.63 & 12.91 & 13.53 & 44.67 & 58.47 & 63.26 & 43.50 & 39.79 & 2.57±0.98\\
      \textbf{GPT2-Large} & 9.31 & 7.78 & 6.69 & 7.78 & 37.12 & 48.24 & 51.84 & \textbf{\textcolor{red}{60.50}} & \textbf{\textcolor{red}{54.90}} & 2.44±0.98\\
      \textbf{GPT2-XL} & 9.53 & 7.90 & 7.51 & 6.35 & 41.00 & 54.15 & 58.93 & 57.50 & 51.96 & 2.44±0.92\\
      \textbf{GS-Small} & 7.56 & 9.54 & 10.03 & 10.76 & 31.84 & 41.83 & 45.42 & 49.00 & 46.41 & 2.74±1.07\\
      \textbf{GS-Medium} & 8.02 & 6.17 & 7.15 & 7.63 & 32.74 & 43.20 & 47.08 & 49.00 & 44.59 & 2.43±0.96\\
      \textbf{GS-Large} & \textbf{\textcolor{red}{6.98}} & 6.12 & \textbf{\textcolor{red}{5.28}} & 6.05 & 31.08 & 41.96 & 46.29 & 55.00 & 51.83 & 2.69±1.01\\
      \textbf{GS-XL} & 7.10 & \textbf{\textcolor{red}{5.86}} & 5.58 & \textbf{\textcolor{red}{4.93}} & 30.18 & 40.55 & 44.66 & 53.00 & 49.65 & 2.45±0.93\\
      \textbf{GSM-Small} & 24.23 & 32.07 & 34.69 & 37.08 & 79.57 & 93.99 & 95.04 & 52.00 & 51.26 & 3.15±0.90\\
      \textbf{GSM-Medium} & 34.67 & 23.77 & 29.83 & 32.16 & \textbf{\textcolor{red}{80.76}} & \textbf{\textcolor{red}{95.20}} & 95.88 & 55.50 & 48.74 & \textbf{\textcolor{red}{3.31±0.81}}\\
      \textbf{GSM-Large} & 22.80 & 18.97 & 15.41 & 18.52 & 75.10 & 91.87 & 93.56 & 54.50 & 50.22 & 3.19±0.91\\
      \textbf{GSM-XL} & 21.78 & 17.45 & 16.36 & 13.50 & 76.36 & 93.43 & 95.15 & 50.50 & 48.93 & \textbf{\textcolor{red}{3.31±0.72}}\\
      \textbf{GPT2-FT} & 17.43 & 12.57 & 35.63 & 85.62 & 51.11 & 67.01 & 71.72 & 39.50 & 38.91 & 2.88±1.01\\
      \textbf{PPLM-Discrim} & 12.83 & 10.52 & 11.25 & 11.88 & 56.01 & 80.50 & 86.29 & 38.00 & 37.80 & 2.88±0.98\\
      \textbf{K2T} & 24.94 & 20.90 & 17.01 & 20.23 & 69.65 & 93.71 & \textbf{\textcolor{red}{96.28}} & 38.00 & 36.47 & 2.97±0.78\\
      \textbf{CTRL} & 10.33 & 8.90 & 8.02 & 7.80 & 45.45 & 65.74 & 71.91 & 52.00 & 48.69 & 3.22±0.81\\
      \bottomrule % <-- Bottomrule here
    \end{tabular}}
  \end{center}
\end{table*}

\end{document}